\documentclass[11pt]{article}

\usepackage[preprint]{acl}

\usepackage{times}
\usepackage{latexsym}

\usepackage[T1]{fontenc}

\usepackage[utf8]{inputenc}

\usepackage{microtype}

\usepackage{inconsolata}

\usepackage{graphicx}

\usepackage{amsmath}
\usepackage{amssymb}
\usepackage{mathtools}
\usepackage{amsthm}
\usepackage{subcaption}
\usepackage{multirow}
\usepackage{CJKutf8}

\usepackage{xcolor}
\definecolor{recoveryblue}{RGB}{33,113,181}
\newcommand{\rec}[1]{\textcolor{recoveryblue}{#1}}

\usepackage{booktabs} 
\title{Word Recovery in Large Language Models Enables Character-Level Tokenization Robustness}

\author{
 \textbf{Zhipeng Yang\textsuperscript{1}},
 \textbf{Shu Yang\textsuperscript{2,3}},
 \textbf{Lijie Hu\textsuperscript{1}},
 \textbf{Di Wang\textsuperscript{2,3, $\dagger$}}
\\
\\
 \textsuperscript{1}Mohamed bin Zayed University of Artificial Intelligence (MBZUAI),\\
 \textsuperscript{2}Provable Responsible AI and Data Analytics (PRADA) Lab,\\
 \textsuperscript{3}King Abdullah University of Science and Technology
}

\begin{document}
\maketitle
\begin{abstract}
Large language models (LLMs) trained with canonical tokenization exhibit surprising robustness to non-canonical inputs such as character-level tokenization, yet the mechanisms underlying this robustness remain unclear. We study this phenomenon through mechanistic interpretability and identify a core process we term \emph{word recovery}. We first introduce a decoding-based method to detect word recovery, showing that hidden states reconstruct canonical word-level token identities from character-level inputs. We then provide causal evidence by removing the corresponding subspace from hidden states, which consistently degrades downstream task performance. Finally, we conduct a fine-grained attention analysis and show that in-group attention among characters belonging to the same canonical token is critical for word recovery: masking such attention in early layers substantially reduces both recovery scores and task performance. Together, our findings provide a mechanistic explanation for tokenization robustness and identify word recovery as a key mechanism enabling LLMs to process character-level inputs.
\end{abstract}

\begingroup
\renewcommand\thefootnote{}\footnotetext{$^\dagger$~Correspondence to: Di Wang <di.wang@kaust.edu.sa>.}
\addtocounter{footnote}{-1}
\endgroup

\begin{figure}[t]
  \centering
  \includegraphics[width=\columnwidth]{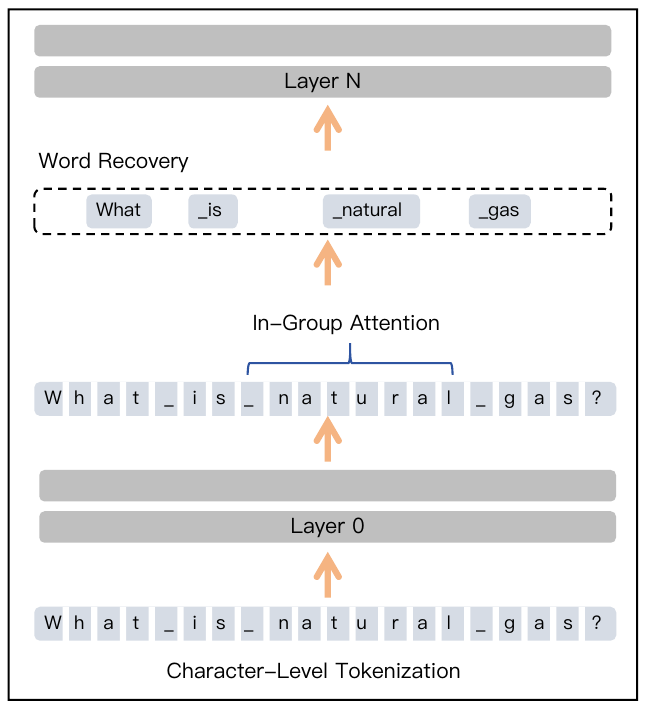}
  \caption{\label{diagram}
Overview of word recovery in large language models under character-level tokenization. A character-level input (e.g., “What is natural gas?”) is processed by the transformer, where early-layer in-group attention aggregates information among characters belonging to the same canonical token. This enables the model to reconstruct word-level representations in hidden states, which are then used for downstream contextual understanding.
}
\end{figure}

\section{Introduction}
Large language models (LLMs) are typically trained and evaluated using a fixed, canonical tokenization scheme, most commonly based on subword methods such as Byte Pair Encoding (BPE), which map raw text into discrete tokens drawn from a learned vocabulary \citep{sennrich2016neural,kudo2018sentencepiece}. This tokenization process is traditionally regarded as a lossy preprocessing step: words are converted into sequences of token indices, and fine-grained character-level information is largely abstracted away before the model’s core computation begins \citep{chai2024tokenization, erdogan2026information, wang2024tokenization}. Under this view, an LLM’s linguistic competence is fundamentally tied to the token units it was trained on, and deviations from the training tokenization should substantially impair performance.

\begin{table*}[t]
\centering
\small
\setlength{\tabcolsep}{3.5pt}

\caption{
\textbf{Performance and word recovery under character-level tokenization.}
For each model and benchmark, we report accuracy under canonical tokenization (\textbf{Canon}), the change in accuracy under character-level tokenization (\textbf{Char $\Delta$}), and the corresponding \textbf{word recovery score} measured from character-level inputs. Since recovery is defined layerwise, we report the maximum recovery score across layers in this table.
}

\begin{tabular}{lcc|c cc|c cc|c}
\toprule
\multirow{2}{*}{\textbf{Benchmark}} 
& \multicolumn{3}{c}{\textbf{Gemma-2-9B-It}} 
& \multicolumn{3}{c}{\textbf{Qwen2.5-7B-Instruct}} 
& \multicolumn{3}{c}{\textbf{Llama-3.2-3B-Instruct}} \\
\cmidrule(lr){2-4} \cmidrule(lr){5-7} \cmidrule(lr){8-10}
& \textbf{Canon}
& \textbf{Char $\Delta$}
& \textbf{Recovery}
& \textbf{Canon}
& \textbf{Char $\Delta$}
& \textbf{Recovery}
& \textbf{Canon}
& \textbf{Char $\Delta$}
& \textbf{Recovery} \\
\midrule

ARC-E        & $96.8$ & $-3.50$ & \rec{96.8} & $95.6$ & $-2.30$ & \rec{53.8} & $88.6$ & $-35.80$ & \rec{69.9} \\
ARC-C        & $90.6$ & $-4.00$ & \rec{96.6} & $91.6$ & $-4.00$ & \rec{55.4} & $76.6$ & $-33.80$ & \rec{71.2} \\
CSQA         & $80.7$ & $-5.90$ & \rec{96.4} & $85.3$ & $-9.10$  & \rec{51.1} & $73.9$ & $-33.40$ & \rec{55.9} \\
OpenbookQA   & $87.8$ & $-7.80$ & \rec{95.7} & $86.6$ & $-8.20$  & \rec{51.6} & $77.0$ & $-30.80$ & \rec{58.6} \\

\bottomrule
\end{tabular}

\label{tab:tokenization_robustness}
\end{table*}

Surprisingly, however, recent studies have shown that LLMs trained with canonical tokenization can still perform competitively when evaluated on non-canonical tokenizations, including character-level sequences and randomly segmented tokenizations that break learned token boundaries \citep{zheng2025broken}. This robustness has been observed across multiple model families and downstream tasks, suggesting that it is not an isolated artifact of a particular architecture or dataset. At the same time, prior work has documented cases where tokenization obscures orthographic or morphological information and leads to brittle behavior \citep{edman2024cute,wang2024tokenization}, motivating a long-standing debate over whether tokenization fundamentally limits language model understanding.

These seemingly conflicting findings challenge the conventional understanding of tokenization as a hard constraint on model capability and raise a central question: how do LLMs internally process and interpret fragmented inputs that deviate from their training tokenization? In particular, it remains unclear whether models directly reason over character-level representations throughout the network, or whether they internally reconstruct higher-level lexical units that resemble canonical tokens. Answering this question requires moving beyond input–output behavior and toward an analysis of the internal representations.

In this work, we address this question through the lens of mechanistic interpretability, aiming to uncover the internal computations that enable LLMs to bridge the gap between character-level inputs and effective language understanding. Concretely, we focus on \emph{character-level tokenization} as a representative and extreme form of non-canonical tokenization, in which the input is decomposed into individual characters and explicit word boundaries are removed (Figure~\ref{diagram}). We identify a core process we term \emph{word recovery}, in which models reconstruct canonical word- or subword-level identities within their hidden states and use these recovered units as intermediates for downstream computation. Our analysis proceeds in three stages. First, we introduce a decoding-based framework that probes hidden states across layers using the model’s input embedding, yielding a layerwise \emph{word recovery score} that quantifies how many canonical tokens become recoverable from character-level representations. Second, we test the functional importance of word recovery via targeted subspace interventions: at selected layers and character spans, we remove the component of the residual stream aligned with specific recovered tokens and measure the resulting degradation in task performance. This intervention reveals a tight correspondence between recovery and causal necessity, with early recovered tokens playing a disproportionate role in enabling accurate predictions. Finally, we perform a fine-grained attention analysis by selectively masking \emph{in-group attention}, attention among characters belonging to the same canonical token. 

Together, these results reveal that robustness to character-level tokenization does not arise from direct character-level reasoning. Instead, LLMs internally reconstruct word-level representations and rely on them as intermediate units for contextual understanding. We show that this robustness arises from an internal, causally relevant \emph{word recovery} mechanism, supported by in-group attention in early layers that aggregates character information into lexical representations. Our work clarifies the internal structure underlying tokenization robustness and contributes to a growing body of mechanistic interpretability research aimed at understanding how modern language models represent and manipulate linguistic information.

\begin{figure*}[t]
    \centering
    \begin{subfigure}[t]{0.32\textwidth}
        \centering
        \includegraphics[width=\textwidth]{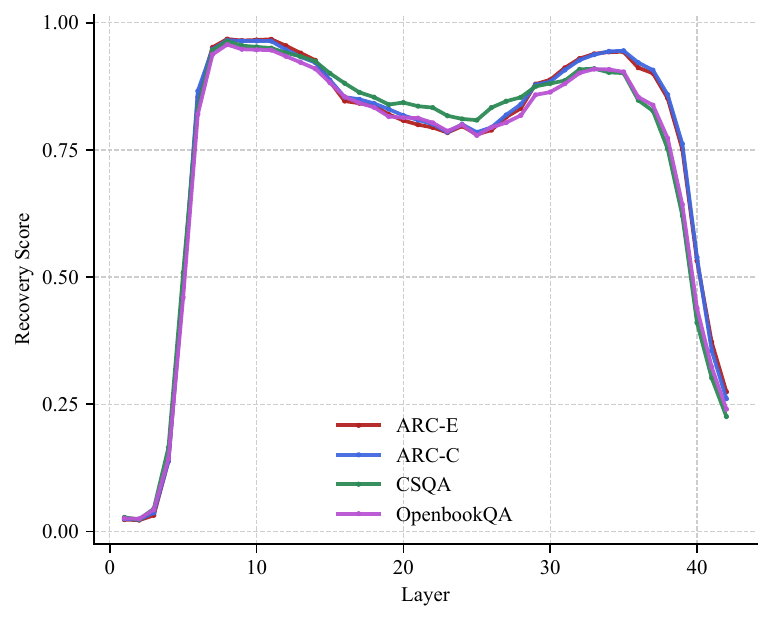}
        \caption{Gemma-2-9b-It}
        \label{fig:recovery:a}
    \end{subfigure}\hfill
    \begin{subfigure}[t]{0.32\textwidth}
        \centering
        \includegraphics[width=\textwidth]{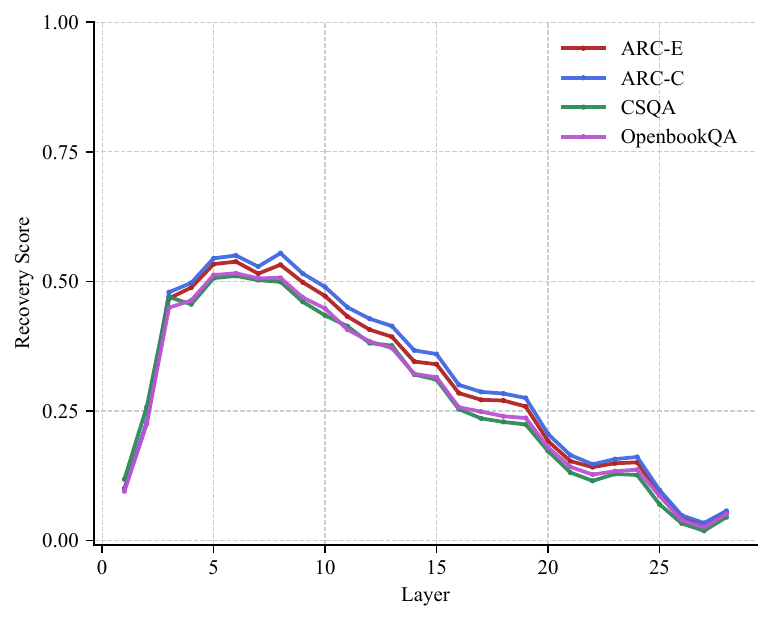}
        \caption{Qwen2.5-7B-Instruct}
        \label{fig:recovery:b}
    \end{subfigure}\hfill
    \begin{subfigure}[t]{0.32\textwidth}
        \centering
        \includegraphics[width=\textwidth]{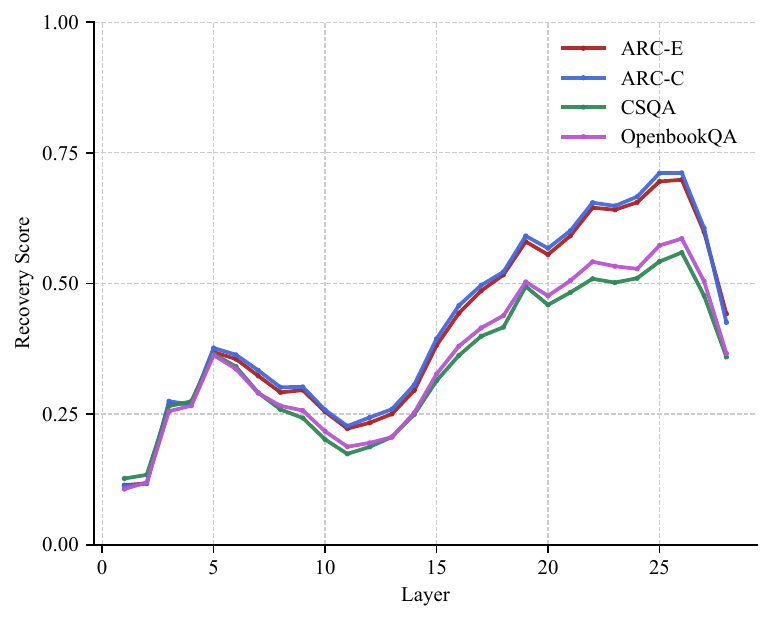}
        \caption{Llama-3.2-3B-Instruct}
        \label{fig:recovery:c}
    \end{subfigure}\hfill

    \caption{
    \textbf{Layerwise word recovery under character-level tokenization.}
    We show the word recovery score across four datasets and three models. Recovery patterns are consistent across datasets within each model, while the layerwise behavior of recovery varies across models.
}

    \label{fig:word_recovery}
\end{figure*}

\section{Background}
\label{sec:background}
\subsection{Character-Level Tokenization}
Large language models (LLMs) are typically trained using a fixed, canonical tokenization scheme, most commonly based on subword methods such as Byte Pair Encoding (BPE). Given an input string $x$, canonical tokenization maps $x$ to a sequence of tokens
\[
x \mapsto (t_1, t_2, \dots, t_N), \quad t_i \in \mathcal{V},
\]
where $\mathcal{V}$ is a learned subword vocabulary and each $t_i$ corresponds to a word or subword unit. The model then operates on the corresponding sequence of token indices. Under this formulation, tokenization is commonly viewed as a preprocessing step that abstracts away internal character structure, since the model does not directly observe the characters composing each token.

In contrast, under character-level tokenization, the same input string $x$ is represented as a sequence of characters
\[
x \mapsto (c_1, c_2, \dots, c_M),
\]
where each $c_j$ is a single character. Each canonical token $t_i$ corresponds to a contiguous span of characters in the original string. We denote this span by
\[
\mathcal{G}(t_i) = \{ c_{s_i}, c_{s_i+1}, \dots, c_{e_i} \},
\]
where the concatenation of characters in $\mathcal{G}(t_i)$ recovers $t_i$. We refer to $\mathcal{G}(t_i)$ as an \emph{in-group}, and each $c_j \in \mathcal{G}(t_i)$ as an \emph{in-group token} with respect to $t_i$. Character tokens belonging to different canonical tokens are referred to as \emph{out-of-group tokens}. This induces a latent grouping structure over character-level inputs, which is not explicitly provided to the model but plays a central role in our subsequent analysis.

\subsection{Models and Datasets}
We conduct our analysis on a diverse set of instruction-tuned large language models to evaluate the generality of the proposed mechanisms across architectures and model scales. Specifically, we study Gemma-2-9B-It \citep{team2024gemma}, Qwen2.5-7B-Instruct \citep{yang2024qwen2}, and Llama-3.2-3B-Instruct \citep{llama32}. We evaluate models on a set of question answering benchmarks, including ARC-Easy (ARC-E), ARC-Challenge (ARC-C), CommonsenseQA (CSQA), and OpenbookQA. These datasets have been used in prior work to study robustness to non-canonical tokenization, showing that models trained with canonical tokenization can retain competitive performance under character-level or randomly segmented inputs \citep{zheng2025broken}. We therefore adopt these benchmarks as a controlled testbed for analyzing internal mechanisms underlying tokenization robustness. For each dataset, we evaluate model performance under both canonical tokenization and character-level tokenization. Summary results are reported in Table~\ref{tab:tokenization_robustness}. All evaluations are conducted without additional fine-tuning, enabling us to isolate how pretrained models internally process character-level inputs.

\begin{figure*}[t]
    \centering
    \begin{subfigure}[t]{0.25\textwidth}
        \centering
        \includegraphics[width=\textwidth]{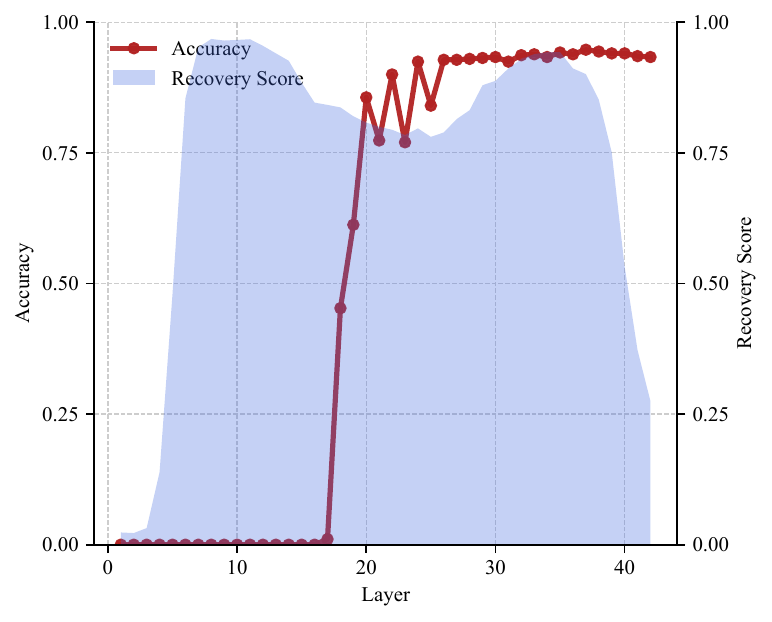}
        \caption{ARC-E}
        \label{fig:causal_gemma:a}
    \end{subfigure}\hfill
    \begin{subfigure}[t]{0.25\textwidth}
        \centering
        \includegraphics[width=\textwidth]{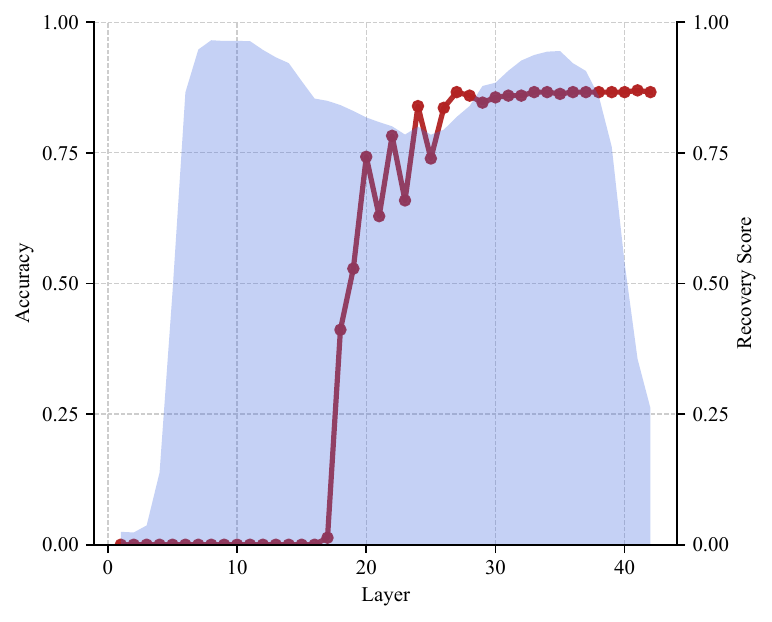}
        \caption{ARC-C}
        \label{fig:causal_gemma:b}
    \end{subfigure}\hfill
    \begin{subfigure}[t]{0.25\textwidth}
        \centering
        \includegraphics[width=\textwidth]{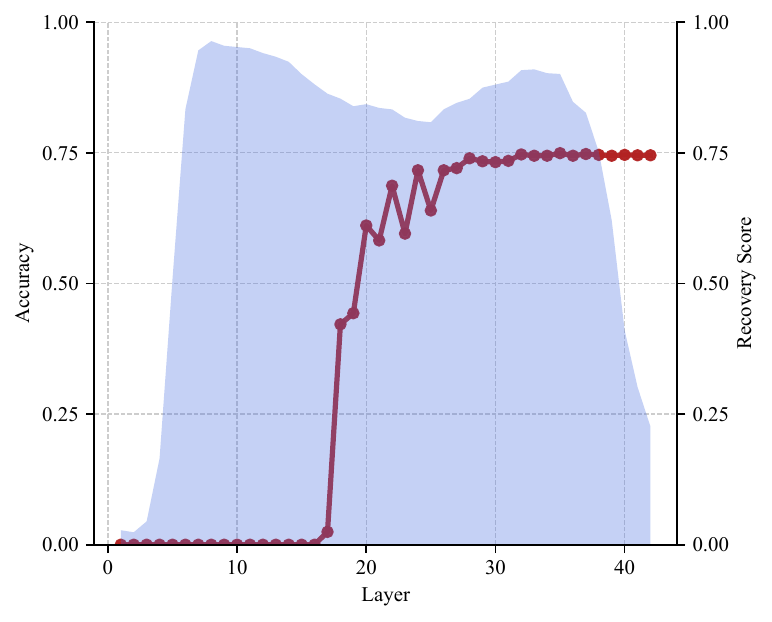}
        \caption{CSQA}
        \label{fig:causal_gemma:c}
    \end{subfigure}\hfill
    \begin{subfigure}[t]{0.25\textwidth}
        \centering
        \includegraphics[width=\textwidth]{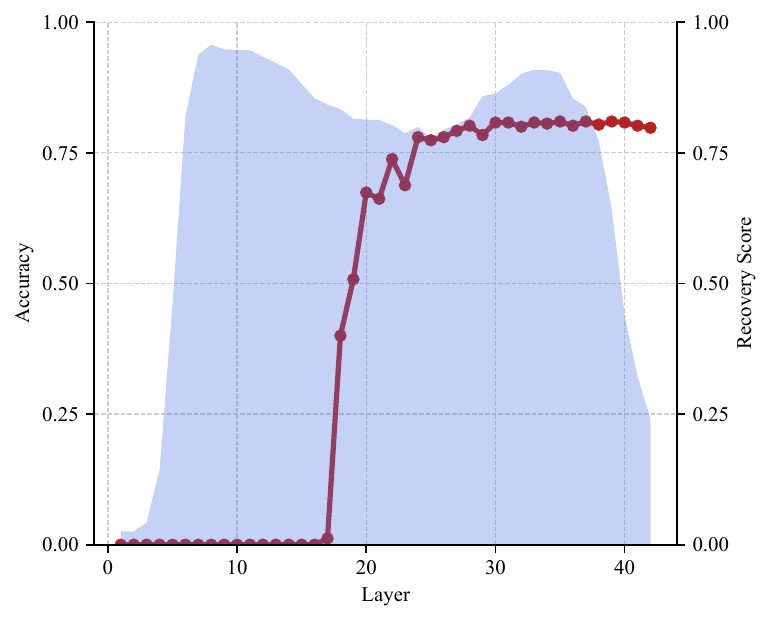}
        \caption{OpenbookQA}
        \label{fig:causal_gemma:d}
    \end{subfigure}\hfill

\caption{
\textbf{Layerwise effects of targeted word-recovery intervention for Gemma-2-9B-It.}
The line plot shows task performance under targeted intervention applied starting from each transformer layer, while the shaded area shows the corresponding word recovery score at the intervention starting layer under character-level tokenization.
}

    \label{fig:causal_gemma}
\end{figure*}

\section{Detecting Word Recovery in Hidden States}
\label{sec:word_recovery}
Although large language models are capable of processing character-level inputs despite being trained with canonical tokenization, it is unclear whether they directly reason over characters or instead reconstruct higher-level lexical units internally. We adopt the hypothesis that LLMs do not natively operate on individual characters as semantic units. Instead, when presented with character-level inputs, they first aggregate character information to recover meaningful word-level or subword-level representations corresponding to canonical tokens, which are then used for downstream computation. Under this hypothesis, robustness to character-level tokenization arises from an internal \emph{word recovery} process rather than direct character-level understanding.

To operationalize this hypothesis, we analyze whether hidden states at different layers encode the original canonical tokens corresponding to the input text. Let $x$ be an input string, canonically tokenized as $(t_1, \dots, t_N)$ under a BPE tokenizer, and represented under character-level tokenization as $(c_1, \dots, c_M)$. For a given layer $\ell$, let $h^{(\ell)}_j$ denote the hidden state corresponding to character token $c_j$. Instead of decoding with the unembedding matrix \citep{logitlen}, we decode hidden states using the model's input embedding matrix, aiming to better reflect the recovery process from character-level inputs to canonical lexical embeddings. Specifically, we compute a distribution over the vocabulary via:
\[
p^{(\ell)}_j = \mathrm{softmax}(W_{\text{embed}} h^{(\ell)}_j),
\]
where $p^{(\ell)}_j \in \mathbb{R}^{|\mathcal{V}|}$. For each position $j$, we extract the top-$K$ predicted token identities according to $p^{(\ell)}_j$. For each input example, we define the set of target canonical tokens $\mathcal{T} = \{ t_1, \dots, t_N \},$ where duplicates are removed to avoid overcounting repeated tokens. Let $\mathcal{P}^{(\ell)}$ denote the union of all tokens appearing in the top-$K$ predictions across the entire character sequence at layer $\ell$:
\[
\mathcal{P}^{(\ell)} = \bigcup_{j=1}^{M} \mathrm{TopK}(p^{(\ell)}_j).
\]
We say that a canonical token $t \in \mathcal{T}$ is \emph{recovered} at layer $\ell$ if $t \in \mathcal{P}^{(\ell)}$. The recovery score at layer $\ell$ is then defined as
\[
R^{(\ell)} = \frac{|\mathcal{T} \cap \mathcal{P}^{(\ell)}|}{|\mathcal{T}|},
\]
which measures the proportion of unique canonical tokens whose identities are recoverable from the layer-$\ell$ hidden states. This set-based formulation captures whether the model has internally reconstructed the lexical identities present in the input, independent of their specific positions in the character sequence. Tracking $R^{(\ell)}$ across layers allows us to characterize when and how word-level representations emerge during processing.

Because our recovery metric aggregates decoded tokens across the entire character sequence, it may overestimate recovery by allowing tokens to be detected outside their original character spans. To address this potential concern, we include an ablation in Appendix~\ref{sec:appendix_ingroup_decoding} that restricts decoding to in-group character ranges, verifying that our core conclusions about the timing and prevalence of word recovery remain unchanged. We further verify that word recovery generalizes beyond character-level tokenization to random segmentation settings, and analyze recovery behavior across different token lengths (Appendix~\ref{sec:appendix_generalization}).

\begin{figure*}[t]
    \centering
    \begin{subfigure}[t]{0.25\textwidth}
        \centering
        \includegraphics[width=\textwidth]{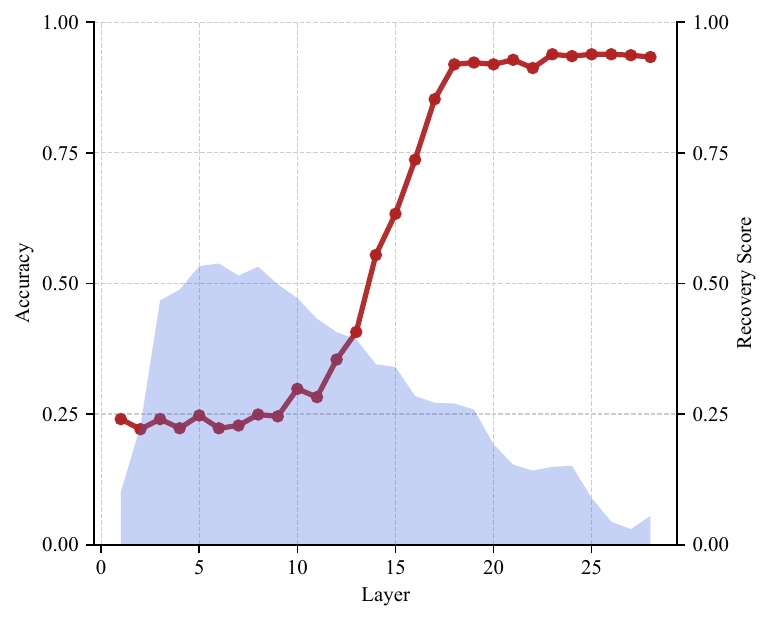}
        \caption{ARC-E}
        \label{fig:causal_qwen:a}
    \end{subfigure}\hfill
    \begin{subfigure}[t]{0.25\textwidth}
        \centering
        \includegraphics[width=\textwidth]{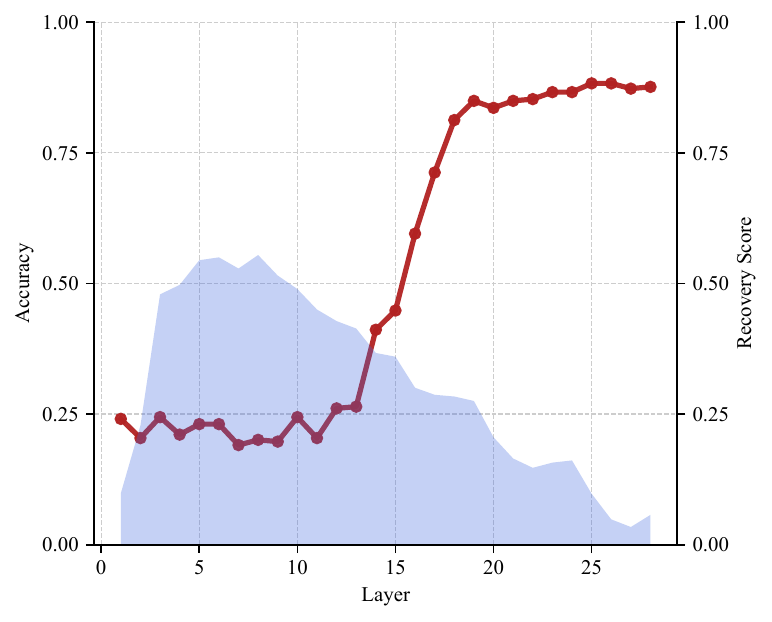}
        \caption{ARC-C}
        \label{fig:causal_qwen:b}
    \end{subfigure}\hfill
    \begin{subfigure}[t]{0.25\textwidth}
        \centering
        \includegraphics[width=\textwidth]{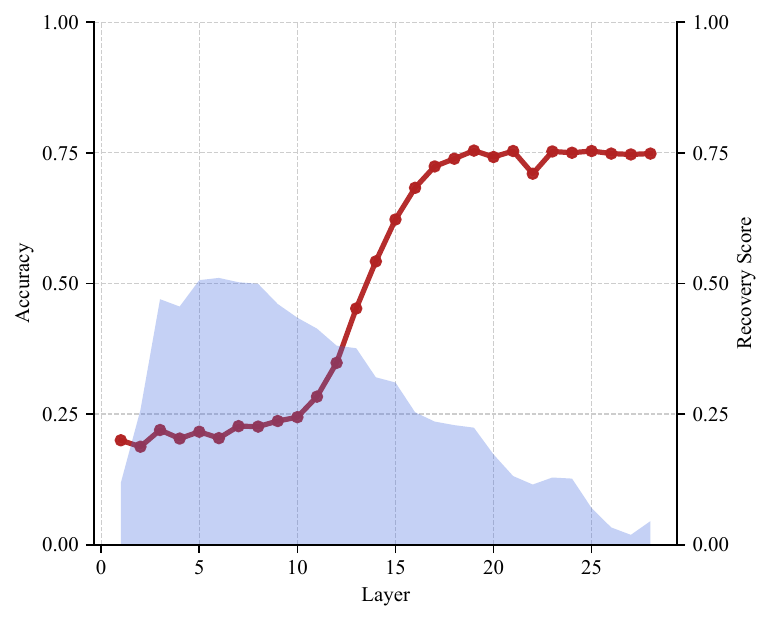}
        \caption{CSQA}
        \label{fig:causal_qwen:c}
    \end{subfigure}\hfill
    \begin{subfigure}[t]{0.25\textwidth}
        \centering
        \includegraphics[width=\textwidth]{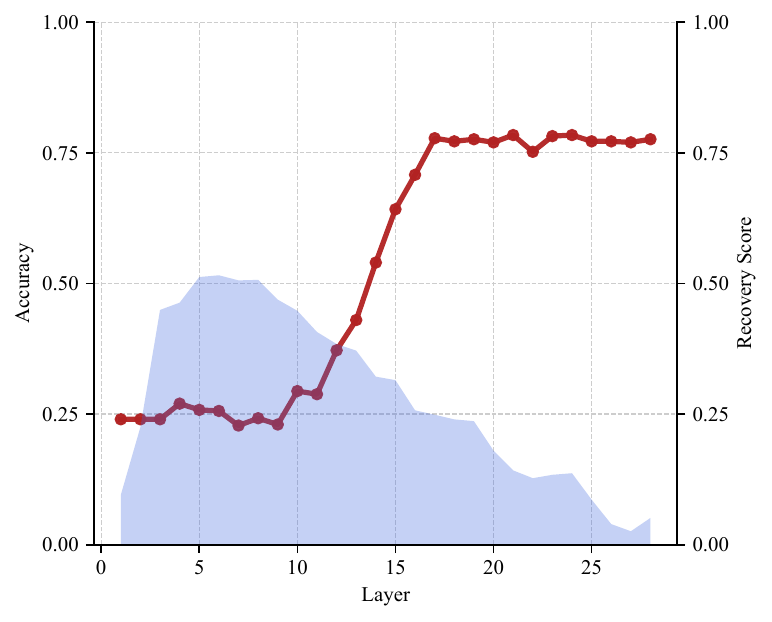}
        \caption{OpenbookQA}
        \label{fig:causal_qwen:d}
    \end{subfigure}\hfill

\caption{
\textbf{Layerwise effects of targeted word-recovery intervention for Qwen2.5-7B-Instruct.}
The line plot shows task performance under targeted intervention applied starting from each transformer layer, while the shaded area shows the corresponding word recovery score at the intervention starting layer under character-level tokenization.
}
    \label{fig:causal_qwen}
\end{figure*}

\begin{figure*}[t]
    \centering
    \begin{subfigure}[t]{0.25\textwidth}
        \centering
        \includegraphics[width=\textwidth]{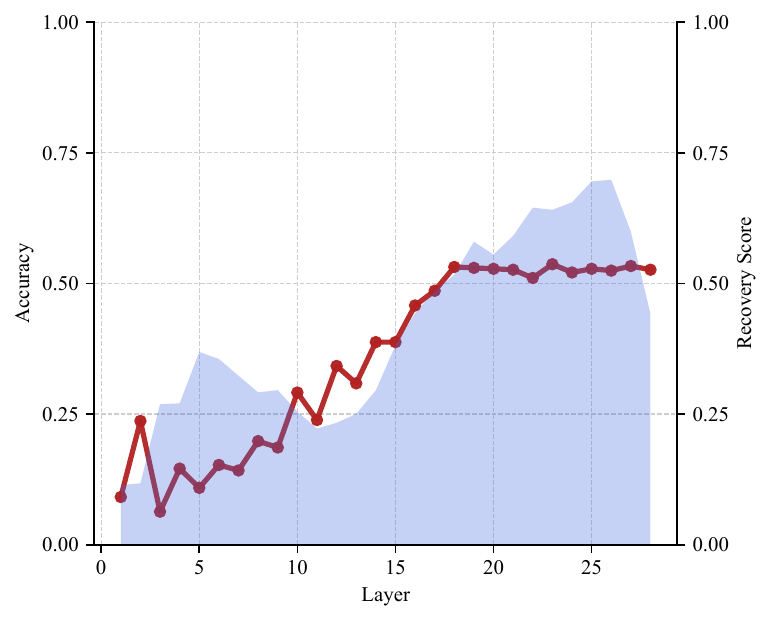}
        \caption{ARC-E}
        \label{fig:causal_llama:a}
    \end{subfigure}\hfill
    \begin{subfigure}[t]{0.25\textwidth}
        \centering
        \includegraphics[width=\textwidth]{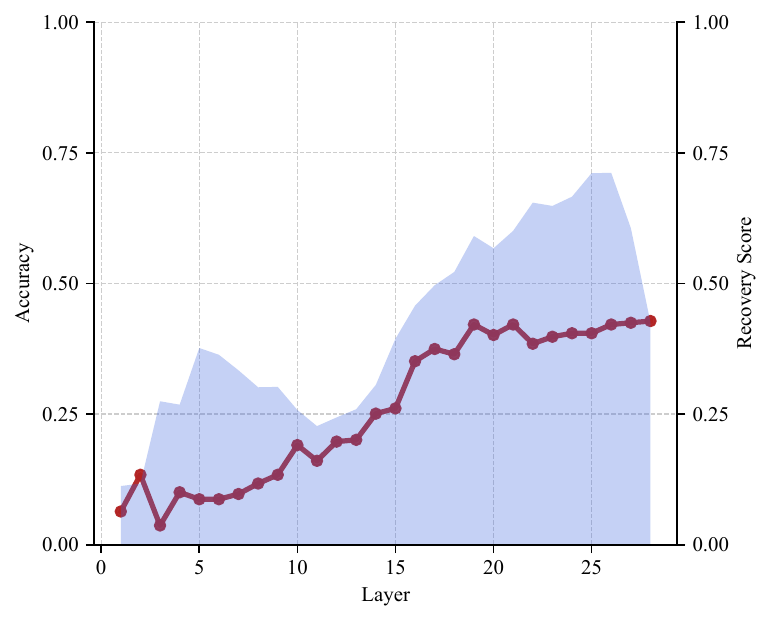}
        \caption{ARC-C}
        \label{fig:causal_llama:b}
    \end{subfigure}\hfill
    \begin{subfigure}[t]{0.25\textwidth}
        \centering
        \includegraphics[width=\textwidth]{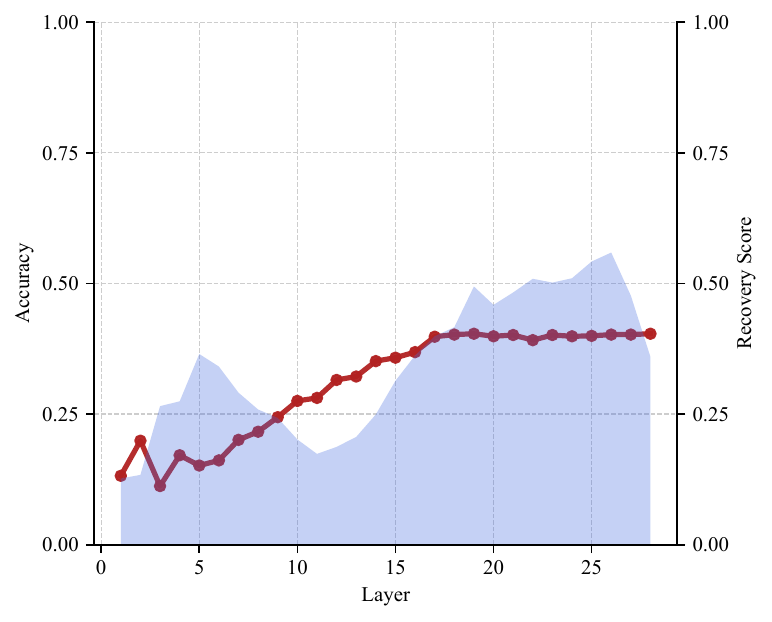}
        \caption{CSQA}
        \label{fig:causal_llama:c}
    \end{subfigure}\hfill
    \begin{subfigure}[t]{0.25\textwidth}
        \centering
        \includegraphics[width=\textwidth]{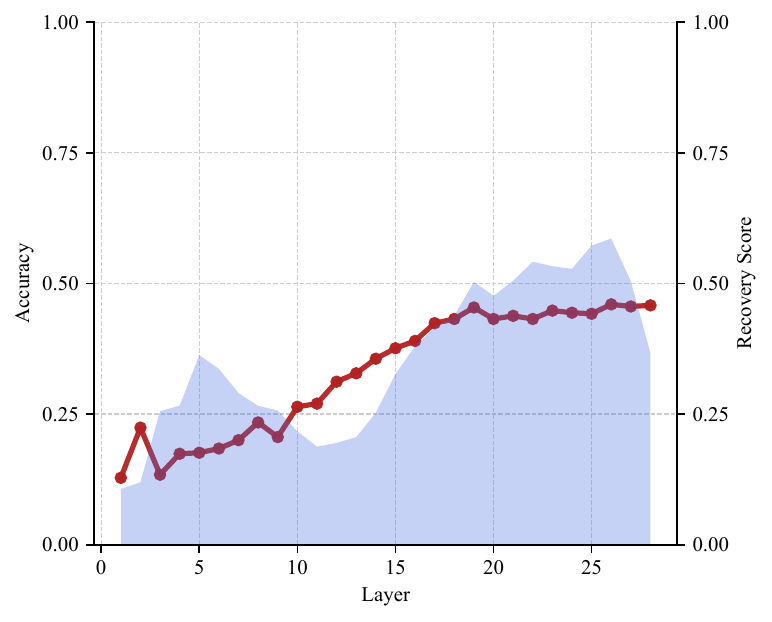}
        \caption{OpenbookQA}
        \label{fig:causal_llama:d}
    \end{subfigure}\hfill

\caption{
\textbf{Layerwise effects of targeted word-recovery intervention for Llama-3.2-3B-Instruct.}
The line plot shows task performance under targeted intervention applied starting from each transformer layer, while the shaded area shows the corresponding word recovery score at the intervention starting layer under character-level tokenization.
}
    \label{fig:causal_llama}
\end{figure*}

\paragraph{Experiment.}
We measure the word recovery score across four datasets (ARC-E, ARC-C, CSQA, and OpenbookQA) and three models (Gemma-2-9B-It, Qwen2.5-7B-Instruct, and Llama-3.2-3B-Instruct) at every transformer layer. Unless otherwise specified, we use $K=5$ for top-$K$ decoding when computing recovery scores (see Appendix~\ref{sec:appendix_k_choice} for an analysis of robustness to the choice of $K$). Table~\ref{tab:tokenization_robustness} summarizes the maximum recovery score achieved across layers for each model and dataset, while Figure~\ref{fig:word_recovery} reports the full layerwise recovery curves. Across all settings, we observe that word recovery is a universal phenomenon: all models are able to recover a substantial fraction of canonical word tokens from character-level inputs in their hidden states. The maximum recovery scores range from 51.1 to 96.8 across models and datasets, indicating that a large portion of lexical information is internally reconstructed despite the absence of canonical tokenization. Moreover, within each model, recovery curves are highly consistent across datasets, suggesting that word recovery reflects a model-level property rather than dataset-specific behavior.

At the same time, the layerwise dynamics of word recovery vary substantially across models. Gemma-2-9B-It recovers most word tokens in early layers, with recovery saturating quickly and remaining high throughout the network. In contrast, Llama-3.2-3B-Instruct exhibits a two-stage recovery pattern: an initial phase in early layers where a subset of word tokens becomes recoverable, followed by a sharp increase in recovery in mid-to-late layers. This difference highlights that, while word recovery is common across models, the internal pathways by which it emerges are model-dependent. To further investigate these disparities, we extend our analysis to additional models with different scales and architectural choices, including Qwen2.5-1.5B/3B, Gemma-2-2B, and Mistral-7B-Instruct-v0.1 (Appendix~\ref{sec:appendix_size}). These additional experiments suggest that early-layer recovery strongly correlates with robustness within a model family, while cross-family differences remain difficult to explain through any single architectural component.

\begin{figure*}[t]
    \centering
    \begin{subfigure}[t]{0.32\textwidth}
        \centering
        \includegraphics[width=\textwidth]{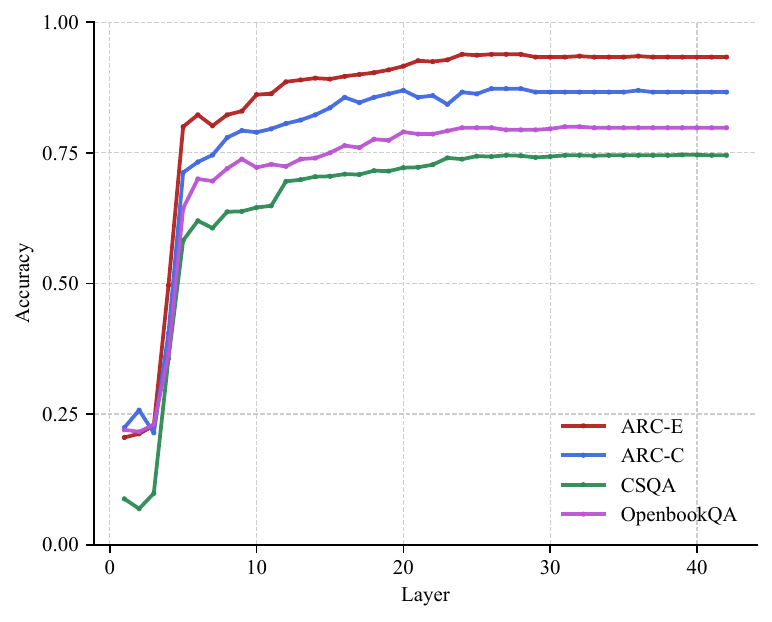}
        \caption{Gemma-2-9b-It}
        \label{fig:masking:a}
    \end{subfigure}\hfill
    \begin{subfigure}[t]{0.32\textwidth}
        \centering
        \includegraphics[width=\textwidth]{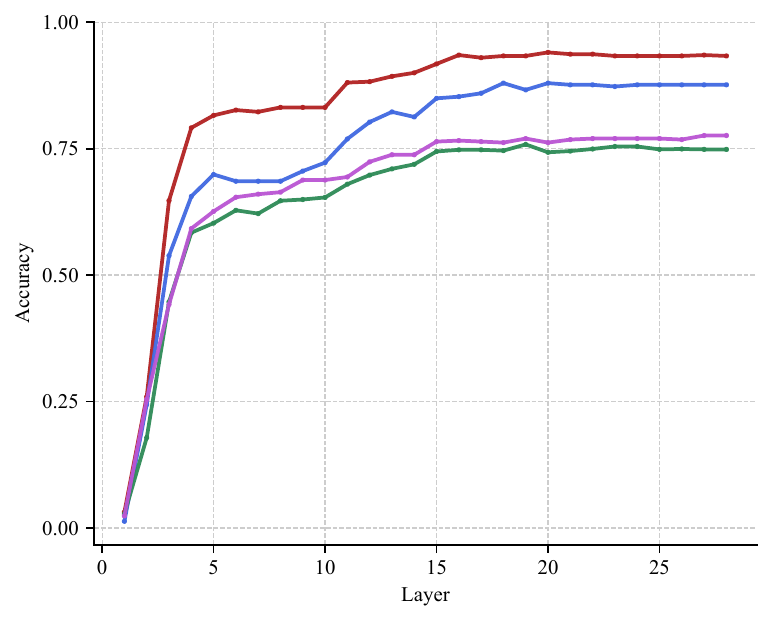}
        \caption{Qwen2.5-7B-Instruct}
        \label{fig:masking:b}
    \end{subfigure}\hfill
    \begin{subfigure}[t]{0.32\textwidth}
        \centering
        \includegraphics[width=\textwidth]{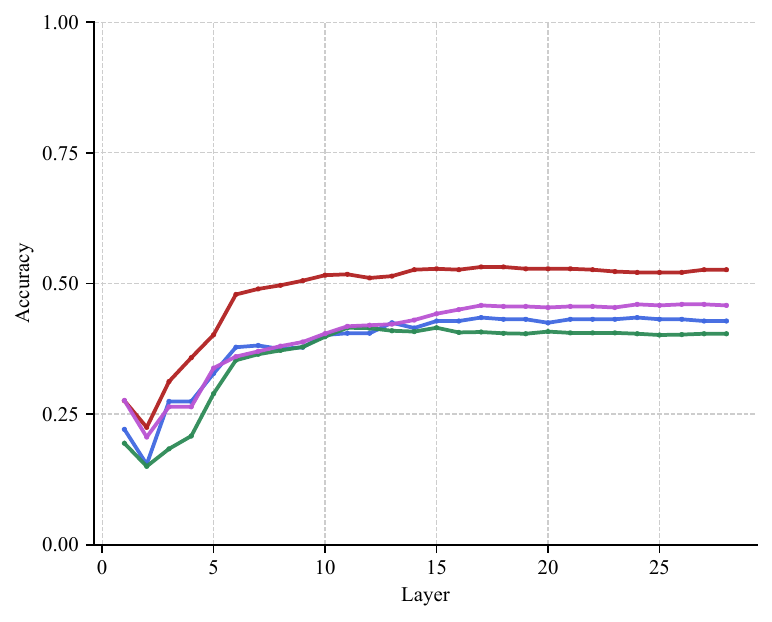}
        \caption{Llama-3.2-3B-Instruct}
        \label{fig:masking:c}
    \end{subfigure}\hfill

\caption{
\textbf{Layerwise effects of in-group attention masking.}
For each model, we plot task accuracy under character-level tokenization when in-group attention (attention among characters belonging to the same canonical token) is masked starting from different transformer layers. Results are shown across four datasets.
}
    \label{fig:masking}
\end{figure*}

\section{Causal Role of Word Recovery}

Decoding analyses reveal that canonical word-level representations are recoverable from hidden states when LLMs process character-level inputs. However, recoverability alone does not imply functional importance: word recovery may emerge as a byproduct of downstream task computation rather than serving as a causal intermediate. To establish a causal role for word recovery, it is therefore necessary to directly intervene on the internal representations corresponding to recovered word tokens and evaluate the impact on model behavior. Our goal is to test whether removing word-level information from hidden states impairs the model’s ability to understand and respond to the input.

We design a targeted intervention that removes the contribution of recovered word tokens from the model’s residual stream. Let $W_{\text{out}} \in \mathbb{R}^{d \times |\mathcal{V}|}$ denote the model’s output embedding matrix, and let $w_t \in \mathbb{R}^d$ be the output embedding corresponding to a canonical token $t$. Following prior work \citep{logitlen}, we treat $w_t$ as a direction in representation space associated with the lexical identity of token $t$. For an input sequence processed under character-level tokenization, consider a canonical token $t$ whose characters occupy positions $[s, e)$ in the character sequence. At a given layer $\ell$, let $h^{(\ell)}_{j} \in \mathbb{R}^d$ denote the residual stream activation at position $j \in [s, e)$. We project these activations onto the token direction $w_t$ and subtract the resulting component:
\[
h^{(\ell)}_{j} \leftarrow h^{(\ell)}_{j} - \langle h^{(\ell)}_{j}, w_t \rangle \, w_t.
\]
This operation removes the contribution of the recovered word token $t$ from the hidden states corresponding to its character span, while preserving orthogonal components of the representation.

\begin{figure*}[t]
    \centering
    \begin{subfigure}[t]{0.32\textwidth}
        \centering
        \includegraphics[width=\textwidth]{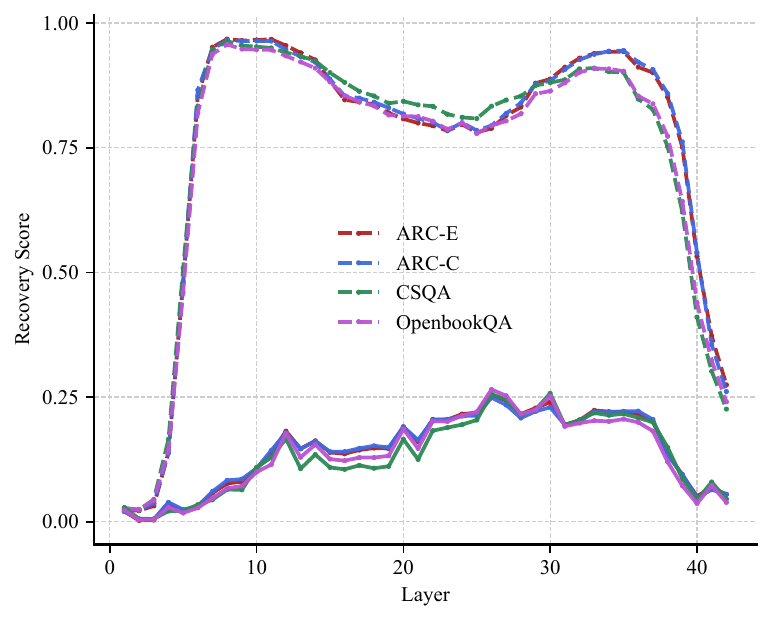}
        \caption{Gemma-2-9b-It}
        \label{fig:mask_decoding:a}
    \end{subfigure}\hfill
    \begin{subfigure}[t]{0.32\textwidth}
        \centering
        \includegraphics[width=\textwidth]{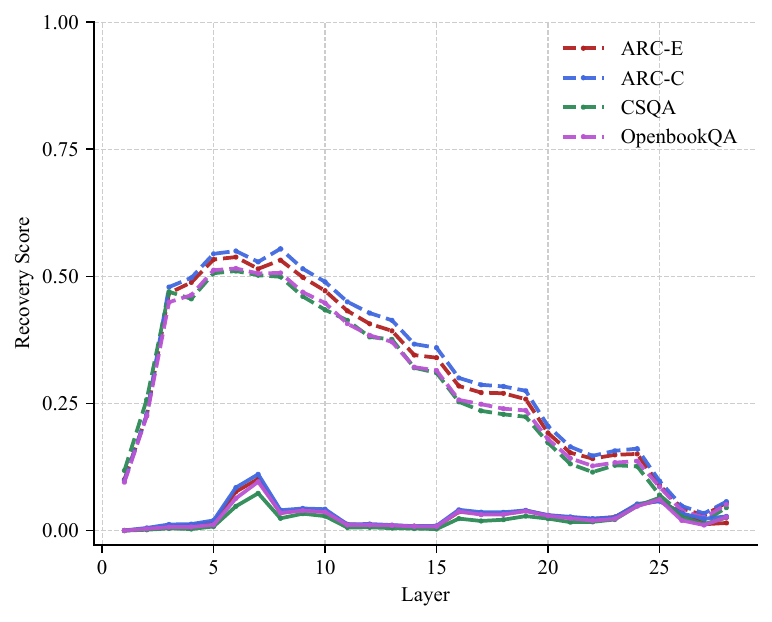}
        \caption{Qwen2.5-7B-Instruct}
        \label{fig:mask_decoding:b}
    \end{subfigure}\hfill
    \begin{subfigure}[t]{0.32\textwidth}
        \centering
        \includegraphics[width=\textwidth]{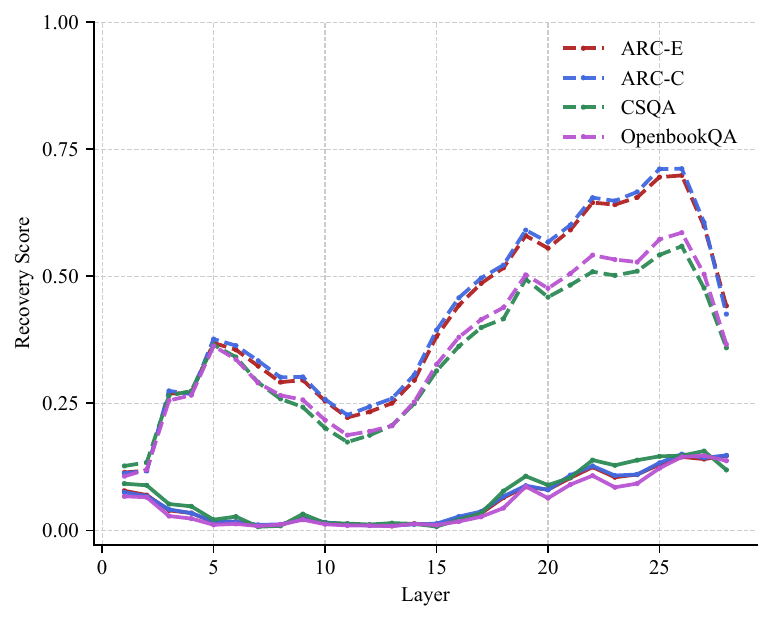}
        \caption{Llama-3.2-3B-Instruct}
        \label{fig:mask_decoding:c}
    \end{subfigure}\hfill

\caption{
\textbf{Early-layer in-group attention supports word recovery.}
We plot word recovery scores across layers when in-group attention is masked in the first five layers, compared to the baseline without masking (dashed). 
}
    \label{fig:mask_decoding}
\end{figure*}

We apply this intervention across a range of layers, starting from a specified layer $\ell_0$ through the final layer, and restrict it to the character positions associated with the target canonical token. By intervening only on a localized subspace and a localized span, we avoid globally perturbing the residual stream and isolate the effect of removing word-level information. This subspace-based removal therefore provides a direct test of whether recovered word representations play a causal role in the model’s contextual understanding.

\paragraph{Experiment.}
We present targeted word-recovery intervention results for three models in Figures~\ref{fig:causal_gemma}, \ref{fig:causal_qwen}, and \ref{fig:causal_llama}. For each benchmark, the line plot reports task performance under targeted intervention applied starting from each transformer layer, while the shaded area shows the corresponding word recovery score at the intervention starting layer under character-level tokenization.

Across all models and datasets, we observe a strong correspondence between word recovery and task performance under intervention. When the intervention is applied starting from early layers where canonical word tokens first become recoverable, task performance drops substantially. This indicates that the recovered word-level representations are functionally necessary for downstream computation. As the intervention starting layer moves deeper into the network, task performance rapidly improves and eventually saturates, suggesting that once recovered word representations have been consumed for contextual understanding, removing them in later layers has little additional effect.

Notably, Llama-3.2-3B-Instruct exhibits a compact, two-stage word recovery process. In early layers, only a small fraction of canonical word tokens (approximately one quarter) are recovered; nevertheless, removing this early recovered subset leads to a large degradation in task performance. In later layers, recovery expands to include most canonical tokens, but interventions applied after this stage no longer substantially affect performance. This pattern suggests that early recovered word tokens play a disproportionate role in enabling context understanding, while word-level representations recovered in later layers are largely redundant for task computation.

Together, these results provide causal evidence that word recovery is not merely an epiphenomenon of model computation, but a necessary intermediate mechanism through which LLMs process character-level inputs.

\section{In-Group Attention Analysis}
Word recovery requires aggregating information distributed across multiple character tokens that together correspond to a single canonical token. In transformer architectures, attention provides the primary mechanism for information integration across positions. We therefore hypothesize that \emph{in-group attention}, attention among characters belonging to the same canonical token, plays a critical role in enabling word recovery. If this hypothesis is correct, selectively disrupting in-group attention should impair the model’s ability to recover word-level representations and, consequently, degrade downstream performance.

Consider an input sequence represented under character-level tokenization as $(c_1, \dots, c_M)$. Let $\mathcal{G}(t_i) = \{c_{s_i}, \dots, c_{e_i}\}$ denote the set of character tokens corresponding to a canonical token $t_i$, as defined in Section~\ref{sec:background}. For a given attention head $h$ at layer $\ell$, let $A^{(\ell,h)} \in \mathbb{R}^{M \times M}$ denote the pre-softmax attention score matrix. We define \emph{in-group attention} as attention from positions within $\mathcal{G}(t_i)$ to other positions in the same group:
\[
A^{(\ell,h)}_{jk} \quad \text{for } j, k \in \mathcal{G}(t_i).
\]
This attention pattern enables characters belonging to the same canonical token to exchange information and form a coherent internal representation. To test the causal role of in-group attention, we design an intervention that selectively suppresses attention within each canonical token group. For each canonical token span $\mathcal{G}(t_i) = [s_i, e_i)$, we modify the attention scores by adding a mask:
\[
A^{(\ell,h)}_{jk} \leftarrow -\infty \quad \text{for all } j, k \in [s_i, e_i),
\]
while leaving all other attention scores unchanged. This masking prevents character tokens within the same group from attending to one another, without affecting attention across different groups or to surrounding context.

We apply this in-group attention masking in a layer-selective manner to distinguish when in-group aggregation is required. Specifically, we consider masking applied from a chosen layer $\ell_0$ through the final layer. By comparing the effects of early-layer and late-layer masking, we can determine whether in-group attention is primarily needed during the initial formation of word-level representations or during later stages of processing. If in-group attention is a necessary mechanism for word recovery, then suppressing it should reduce both the recoverability of canonical tokens in hidden states and the model’s downstream performance. 

\paragraph{Experiment.}
We evaluate the role of in-group attention by measuring task performance under layerwise in-group attention masking, as shown in Figure~\ref{fig:masking}. For each model, we mask in-group attention starting from different transformer layers and report accuracy under character-level tokenization across four datasets. We find that masking in-group attention in early layers leads to a substantial drop in task performance, while masking in later layers has a significantly smaller effect. This pattern confirms the importance of information exchange among in-group tokens during early stages of processing.

To directly assess the impact of in-group attention on word recovery, we further measure recovery scores under early-layer masking. Based on the in-group masking dynamics in Figure~\ref{fig:masking}, we mask in-group attention in the first five layers and compute the resulting word recovery score across all layers. Figure~\ref{fig:mask_decoding} plots recovery scores under masking (solid lines) alongside the baseline without masking (dashed lines). Across all models and datasets, early-layer in-group attention masking leads to a consistent reduction in word recovery scores, indicating that in-group attention in early layers is critical for enabling the word recovery process. We further provide a qualitative example showing that in-group attention masking specifically disrupts recovery of the target word identity while preserving overall model functionality (Appendix~\ref{sec:appendix_qualitative_masking}).

\section{Related Works}

\paragraph{Tokenization Robustness.}
Recent work has shown that large language models exhibit surprising robustness to non-canonical tokenization schemes. In particular, \citet{zheng2025broken} show that instruction-tuned LLMs retain strong performance under character-level and randomly segmented tokenizations. These findings challenge the view that subword tokenization forms a strict bottleneck for language understanding.

A separate line of work investigates how LLMs internally represent lexical information. Prior studies show that token embeddings retain substantial character-level and orthographic structure, enabling sensitivity to spelling patterns even when character composition is not explicitly represented in the input sequence \citep{kaushal2022tokens, itzhak2022models,cosma2025strawberry}. Other work argues that LLMs implicitly construct higher-level lexical representations from fragmented token inputs. For example, \citet{feucht2024token} study implicit vocabulary items through token erasure analysis, while \citet{kaplantokens} analyze how models merge subword representations into coherent word-level concepts within canonical tokenization settings.

Our work connects these two directions by studying the mechanism underlying robustness to non-canonical tokenization. Unlike prior work on lexical reconstruction, which primarily operates within canonical subword tokenization, we focus on character-level tokenization where token boundaries are entirely removed and must be inferred internally. Moreover, while prior studies mainly provide observational evidence for lexical representation formation, we provide causal evidence by directly intervening on recovered word-level representations.

\paragraph{Mechanistic Interpretability.}
Mechanistic interpretability seeks to explain the behavior of large language models by identifying the internal computations and representations that give rise to their outputs \cite{elhage2021mathematical}. A substantial body of work aims to uncover structured transformer circuits—compositions of attention heads and neurons that implement specific algorithmic functions \cite{olsson2022context, gouldsuccessor, wanginterpretability, marks2024sparse, zhang2025mechanistic, cui2025compositional,zhang2025eap,wang2025pahq}. Probing methods such as the LogitLens decode hidden states into the output vocabulary\cite{logitlen}. These techniques have been extended to track hidden-state dynamics and reasoning trajectories in LLMs, revealing structured internal computation across depth \cite{dar2023analyzing, halawioverthinking, merullo2024language, wiegreffe2024answer, yang2025internal,chen2025repreguard}. 

Causal intervention provides a complementary perspective by testing the functional necessity of internal components. Prior work intervenes on neurons, attention heads, or representation subspaces to quantify their causal contribution to model predictions \cite{vig2020investigating, meng2022locating, geva2023dissecting, zhang2025understanding, zhanglocate, jiang2025msrs,wang2025truth,dong2025understanding,su2025understanding}. Finally, recent research addresses the challenge of superposition, where multiple concepts are encoded within shared activations—using sparse autoencoders and related techniques to extract disentangled, interpretable features from high-dimensional representations \cite{elhage2022toy, scherlis2022polysemanticity, gao2024scaling, saeclaude, ferrando2024know,yu2025pixel,yao2025understanding}.

\section{Conclusion}

We study why large language models remain robust under character-level tokenization despite being trained with canonical subword tokenization. Through mechanistic interpretability analyses, we identify a word recovery process in which models internally reconstruct lexical representations from fragmented character inputs. We further show that these recovered representations are causally important for downstream reasoning and are supported by early-layer in-group attention. Beyond explaining tokenization robustness, our findings suggest that LLMs operate on lexical units that are not strictly tied to tokenizer boundaries. Prior work has also shown that non-canonical tokenizations can improve tasks such as string manipulation and code understanding. Our results may support future train-free methods that leverage multiple tokenization schemes for improved downstream performance.

\section*{Limitations}
This work focuses primarily on character-level tokenization as an extreme form of non-canonical tokenization. Although we additionally examine random segmentation tokenization, our analysis does not cover the full range of possible tokenization schemes, such as multilingual, morphology-aware, or byte-level tokenizations. Future work could investigate whether the proposed word recovery mechanism generalizes to these settings.
Our work identifies word recovery as a causally important intermediate process, but does not fully characterize the exact circuits, attention heads, or neurons responsible for implementing this mechanism. Identifying these components remains an important direction for future mechanistic interpretability research.


\bibliography{custom}

@article{su2025understanding,
  title={Understanding how value neurons shape the generation of specified values in llms},
  author={Su, Yi and Zhang, Jiayi and Yang, Shu and Wang, Xinhai and Hu, Lijie and Wang, Di},
  journal={arXiv preprint arXiv:2505.17712},
  year={2025}
}

@article{chen2025repreguard,
  title={Repreguard: Detecting llm-generated text by revealing hidden representation patterns},
  author={Chen, Xin and Wu, Junchao and Yang, Shu and Zhan, Runzhe and Wu, Zeyu and Luo, Ziyang and Wang, Di and Yang, Min and Chao, Lidia S and Wong, Derek F},
  journal={Transactions of the Association for Computational Linguistics},
  volume={13},
  pages={1812--1831},
  year={2025},
  publisher={MIT Press 255 Main Street, 9th Floor, Cambridge, Massachusetts 02142, USA~…}
}

@article{dong2025understanding,
  title={Understanding and Mitigating Cross-lingual Privacy Leakage via Language-specific and Universal Privacy Neurons},
  author={Dong, Wenshuo and Yang, Qingsong and Yang, Shu and Hu, Lijie and Ding, Meng and Lin, Wanyu and Zheng, Tianhang and Wang, Di},
  journal={arXiv preprint arXiv:2506.00759},
  year={2025}
}

@article{yao2025understanding,
  title={Understanding the repeat curse in large language models from a feature perspective},
  author={Yao, Junchi and Yang, Shu and Xu, Jianhua and Hu, Lijie and Li, Mengdi and Wang, Di},
  journal={arXiv preprint arXiv:2504.14218},
  year={2025}
}

@article{wang2025truth,
  title={When truth is overridden: Uncovering the internal origins of sycophancy in large language models},
  author={Wang, Keyu and Li, Jin and Yang, Shu and Zhang, Zhuoran and Wang, Di},
  journal={arXiv preprint arXiv:2508.02087},
  year={2025}
}

@article{zhang2025understanding,
  title={Understanding and mitigating political stance cross-topic generalization in large language models},
  author={Zhang, Jiayi and Yang, Shu and Wu, Junchao and Wong, Derek F and Wang, Di},
  journal={arXiv preprint arXiv:2508.02360},
  year={2025}
}

@article{yu2025pixel,
  title={Pixel: Adaptive steering via position-wise injection with exact estimated levels under subspace calibration},
  author={Yu, Manjiang and Li, Hongji and Singh, Priyanka and Li, Xue and Wang, Di and Hu, Lijie},
  journal={arXiv preprint arXiv:2510.10205},
  year={2025}
}

@article{wang2025pahq,
  title={PAHQ: Accelerating Automated Circuit Discovery through Mixed-Precision Inference Optimization},
  author={Wang, Xinhai and Yang, Shu and Wang, Liangyu and Zhang, Lin and Xie, Huanyi and Hu, Lijie and Wang, Di},
  journal={arXiv preprint arXiv:2510.23264},
  year={2025}
}

@article{zhang2025eap,
  title={Eap-gp: Mitigating saturation effect in gradient-based automated circuit identification},
  author={Zhang, Lin and Dong, Wenshuo and Zhang, Zhuoran and Yang, Shu and Hu, Lijie and Liu, Ninghao and Zhou, Pan and Wang, Di},
  journal={arXiv preprint arXiv:2502.06852},
  year={2025}
}

@inproceedings{edman2024cute,
  title={CUTE: Measuring LLMs’ Understanding of Their Tokens},
  author={Edman, Lukas and Schmid, Helmut and Fraser, Alexander},
  booktitle={Proceedings of the 2024 Conference on Empirical Methods in Natural Language Processing},
  pages={3017--3026},
  year={2024}
}

@inproceedings{chai2024tokenization,
  title={Tokenization Falling Short: On Subword Robustness in Large Language Models},
  author={Chai, Yekun and Fang, Yewei and Peng, Qiwei and Li, Xuhong},
  booktitle={Findings of the Association for Computational Linguistics: EMNLP 2024},
  pages={1582--1599},
  year={2024}
}

@article{wang2024tokenization,
  title={Tokenization matters! degrading large language models through challenging their tokenization},
  author={Wang, Dixuan and Li, Yanda and Jiang, Junyuan and Ding, Zepeng and Luo, Ziqin and Jiang, Guochao and Liang, Jiaqing and Yang, Deqing},
  journal={arXiv preprint arXiv:2405.17067},
  year={2024}
}

@inproceedings{kaushal2022tokens,
  title={What do tokens know about their characters and how do they know it?},
  author={Kaushal, Ayush and Mahowald, Kyle},
  booktitle={Proceedings of the 2022 Conference of the North American Chapter of the Association for Computational Linguistics: Human Language Technologies},
  pages={2487--2507},
  year={2022}
}

@inproceedings{itzhak2022models,
  title={Models in a spelling bee: Language models implicitly learn the character composition of tokens},
  author={Itzhak, Itay and Levy, Omer},
  booktitle={Proceedings of the 2022 Conference of the North American Chapter of the Association for Computational Linguistics: Human Language Technologies},
  pages={5061--5068},
  year={2022}
}

@inproceedings{feucht2024token,
  title={Token erasure as a footprint of implicit vocabulary items in LLMs},
  author={Feucht, Sheridan and Atkinson, David and Wallace, Byron C and Bau, David},
  booktitle={Proceedings of the 2024 Conference on Empirical Methods in Natural Language Processing},
  pages={9727--9739},
  year={2024}
}

@inproceedings{kaplantokens,
  title={From Tokens to Words: On the Inner Lexicon of LLMs},
  author={Kaplan, Guy and Oren, Matanel and Reif, Yuval and Schwartz, Roy},
  booktitle={The Thirteenth International Conference on Learning Representations},
  year={2025}
}

@article{elhage2021mathematical,
  title={A mathematical framework for transformer circuits},
  author={Elhage, Nelson and Nanda, Neel and Olsson, Catherine and Henighan, Tom and Joseph, Nicholas and Mann, Ben and Askell, Amanda and Bai, Yuntao and Chen, Anna and Conerly, Tom and others},
  journal={Transformer Circuits Thread},
  volume={1},
  number={1},
  pages={12},
  year={2021}
}

@article{olsson2022context,
  title={In-context learning and induction heads},
  author={Olsson, Catherine and Elhage, Nelson and Nanda, Neel and Joseph, Nicholas and DasSarma, Nova and Henighan, Tom and Mann, Ben and Askell, Amanda and Bai, Yuntao and Chen, Anna and others},
  journal={arXiv preprint arXiv:2209.11895},
  year={2022}
}

@inproceedings{wanginterpretability,
  title={Interpretability in the Wild: a Circuit for Indirect Object Identification in GPT-2 Small},
  author={Wang, Kevin Ro and Variengien, Alexandre and Conmy, Arthur and Shlegeris, Buck and Steinhardt, Jacob},
  booktitle={The Eleventh International Conference on Learning Representations},
  year={2023}
}

@inproceedings{gouldsuccessor,
  title={Successor Heads: Recurring, Interpretable Attention Heads In The Wild},
  author={Gould, Rhys and Ong, Euan and Ogden, George and Conmy, Arthur},
  booktitle={The Twelfth International Conference on Learning Representations},
  year={2024}
}

@article{marks2024sparse,
  title={Sparse feature circuits: Discovering and editing interpretable causal graphs in language models},
  author={Marks, Samuel and Rager, Can and Michaud, Eric J and Belinkov, Yonatan and Bau, David and Mueller, Aaron},
  journal={arXiv preprint arXiv:2403.19647},
  year={2024}
}

@Misc{logitlen,
howpublished = {\url{https://www.lesswrong.com/posts/AcKRB8wDpdaN6v6ru/interpreting-gpt-the-logit-lens}},
title = {Interpreting gpt: the logit lens},
year = {2020},
author={Nostalgebraist.}
}

@inproceedings{dar2023analyzing,
  title={Analyzing Transformers in Embedding Space},
  author={Dar, Guy and Geva, Mor and Gupta, Ankit and Berant, Jonathan},
  booktitle={Proceedings of the 61st Annual Meeting of the Association for Computational Linguistics (Volume 1: Long Papers)},
  pages={16124--16170},
  year={2023}
}

@inproceedings{halawioverthinking,
  title={Overthinking the Truth: Understanding how Language Models Process False Demonstrations},
  author={Halawi, Danny and Denain, Jean-Stanislas and Steinhardt, Jacob},
  booktitle={The Twelfth International Conference on Learning Representations},
  year={2024}
}

@inproceedings{merullo2024language,
  title={Language Models Implement Simple Word2Vec-style Vector Arithmetic},
  author={Merullo, Jack and Eickhoff, Carsten and Pavlick, Ellie},
  booktitle={Proceedings of the 2024 Conference of the North American Chapter of the Association for Computational Linguistics: Human Language Technologies (Volume 1: Long Papers)},
  pages={5030--5047},
  year={2024}
}

@article{wiegreffe2024answer,
  title={Answer, assemble, ace: Understanding how transformers answer multiple choice questions},
  author={Wiegreffe, Sarah and Tafjord, Oyvind and Belinkov, Yonatan and Hajishirzi, Hannaneh and Sabharwal, Ashish},
  journal={arXiv preprint arXiv:2407.15018},
  year={2024}
}

@article{vig2020investigating,
  title={Investigating gender bias in language models using causal mediation analysis},
  author={Vig, Jesse and Gehrmann, Sebastian and Belinkov, Yonatan and Qian, Sharon and Nevo, Daniel and Singer, Yaron and Shieber, Stuart},
  journal={Advances in neural information processing systems},
  volume={33},
  pages={12388--12401},
  year={2020}
}

@article{meng2022locating,
  title={Locating and editing factual associations in gpt},
  author={Meng, Kevin and Bau, David and Andonian, Alex and Belinkov, Yonatan},
  journal={Advances in neural information processing systems},
  volume={35},
  pages={17359--17372},
  year={2022}
}

@inproceedings{geva2023dissecting,
  title={Dissecting Recall of Factual Associations in Auto-Regressive Language Models},
  author={Geva, Mor and Bastings, Jasmijn and Filippova, Katja and Globerson, Amir},
  booktitle={Proceedings of the 2023 Conference on Empirical Methods in Natural Language Processing},
  pages={12216--12235},
  year={2023}
}

@article{elhage2022toy,
  title={Toy models of superposition},
  author={Elhage, Nelson and Hume, Tristan and Olsson, Catherine and Schiefer, Nicholas and Henighan, Tom and Kravec, Shauna and Hatfield-Dodds, Zac and Lasenby, Robert and Drain, Dawn and Chen, Carol and others},
  journal={arXiv preprint arXiv:2209.10652},
  year={2022}
}

@article{scherlis2022polysemanticity,
  title={Polysemanticity and capacity in neural networks},
  author={Scherlis, Adam and Sachan, Kshitij and Jermyn, Adam S and Benton, Joe and Shlegeris, Buck},
  journal={arXiv preprint arXiv:2210.01892},
  year={2022}
}

@article{gao2024scaling,
  title={Scaling and evaluating sparse autoencoders},
  author={Gao, Leo and la Tour, Tom Dupr{\'e} and Tillman, Henk and Goh, Gabriel and Troll, Rajan and Radford, Alec and Sutskever, Ilya and Leike, Jan and Wu, Jeffrey},
  journal={arXiv preprint arXiv:2406.04093},
  year={2024}
}

@inproceedings{saeclaude,
title = {Scaling monosemanticity: Extracting interpretable features from claude 3 sonnet.},
year = {2024},
author={Anthropic.},
booktitle={Transformer Circuits Thread}
}

@article{ferrando2024know,
  title={Do I Know This Entity? Knowledge Awareness and Hallucinations in Language Models},
  author={Ferrando, Javier and Obeso, Oscar and Rajamanoharan, Senthooran and Nanda, Neel},
  journal={arXiv preprint arXiv:2411.14257},
  year={2024}
}

@inproceedings{sennrich2016neural,
  title={Neural machine translation of rare words with subword units},
  author={Sennrich, Rico and Haddow, Barry and Birch, Alexandra},
  booktitle={Proceedings of the 54th annual meeting of the association for computational linguistics (volume 1: long papers)},
  pages={1715--1725},
  year={2016}
}

@article{kudo2018sentencepiece,
  title={SentencePiece: A simple and language independent subword tokenizer and detokenizer for Neural Text Processing},
  author={Kudo, Taku and Richardson, John},
  journal={EMNLP 2018},
  pages={66},
  year={2018}
}

@article{erdogan2026information,
  title={An Information-Theoretic Perspective on LLM Tokenizers},
  author={Erdogan, Mete and Gorle, Abhiram and Chandak, Shubham and Pilanci, Mert and Weissman, Tsachy},
  journal={arXiv preprint arXiv:2601.09039},
  year={2026}
}

@article{zheng2025broken,
  title={Broken Tokens? Your Language Model can Secretly Handle Non-Canonical Tokenizations},
  author={Zheng, Brian Siyuan and Liu, Alisa and Ahia, Orevaoghene and Hayase, Jonathan and Choi, Yejin and Smith, Noah A},
  journal={arXiv preprint arXiv:2506.19004},
  year={2025}
}

@article{team2024gemma,
  title={Gemma 2: Improving open language models at a practical size},
  author={Team, Gemma and Riviere, Morgane and Pathak, Shreya and Sessa, Pier Giuseppe and Hardin, Cassidy and Bhupatiraju, Surya and Hussenot, L{\'e}onard and Mesnard, Thomas and Shahriari, Bobak and Ram{\'e}, Alexandre and others},
  journal={arXiv preprint arXiv:2408.00118},
  year={2024}
}

@article{yang2024qwen2,
  title={Qwen2. 5 technical report},
  author={Yang, An and Yang, Baosong and Zhang, Beichen and Hui, Binyuan and Zheng, Bo and Yu, Bowen and Li, Chengyuan and Liu, Dayiheng and Huang, Fei and Wei, Haoran and others},
  journal={arXiv preprint arXiv:2412.15115},
  year={2024}
}

@Misc{llama32,
title = {Llama 3.2: Revolutionizing edge AI and vision with open, customizable models},
year = {2024},
author={AI Meta.}
}

@inproceedings{yang2025internal,
  title={Internal Chain-of-Thought: Empirical Evidence for Layer-wise Subtask Scheduling in LLMs},
  author={Yang, Zhipeng and Li, Junzhuo and Xia, Siyu and Hu, Xuming},
  booktitle={Proceedings of the 2025 Conference on Empirical Methods in Natural Language Processing},
  pages={22547--22575},
  year={2025}
}

@inproceedings{zhanglocate,
  title={Locate-then-edit for Multi-hop Factual Recall under Knowledge Editing},
  author={Zhang, Zhuoran and Li, Yongxiang and Kan, Zijian and Cheng, Keyuan and Hu, Lijie and Wang, Di},
  booktitle={Forty-second International Conference on Machine Learning},
  year={2025}
}

@inproceedings{zhang2025mechanistic,
  title={Mechanistic Unveiling of Transformer Circuits: Self-Influence as a Key to Model Reasoning},
  author={Zhang, Lin and Hu, Lijie and Wang, Di},
  booktitle={Findings of the Association for Computational Linguistics: NAACL 2025},
  pages={1387--1404},
  year={2025}
}

@article{cui2025compositional,
  title={The Compositional Architecture of Regret in Large Language Models},
  author={Cui, Xiangxiang and Yang, Shu and Huang, Tianjin and Lin, Wanyu and Hu, Lijie and Wang, Di},
  journal={arXiv preprint arXiv:2506.15617},
  year={2025}
}

@article{jiang2025msrs,
  title={Msrs: Adaptive multi-subspace representation steering for attribute alignment in large language models},
  author={Jiang, Xinyan and Zhang, Lin and Zhang, Jiayi and Yang, Qingsong and Hu, Guimin and Wang, Di and Hu, Lijie},
  journal={arXiv preprint arXiv:2508.10599},
  year={2025}
}

@article{cosma2025strawberry,
  title={The Strawberry Problem: Emergence of Character-level Understanding in Tokenized Language Models},
  author={Cosma, Adrian and Ruseti, Stefan and Radoi, Emilian and Dascalu, Mihai},
  journal={arXiv preprint arXiv:2505.14172},
  year={2025}
}

\appendix
\clearpage

\section{Generalization Beyond Character-Level Tokenization and Token-Type Analysis}
\label{sec:appendix_generalization}

\paragraph{Random segmentation tokenization.}
The main paper focuses on character-level tokenization as an extreme form of non-canonical tokenization where token boundaries are entirely removed. To evaluate whether word recovery generalizes beyond this setting, we additionally study random segmentation tokenization, where canonical word tokens are randomly split into valid vocabulary fragments. Compared to character-level tokenization, this setting preserves partial subword structure while still disrupting the canonical token boundaries seen during training.

\begin{figure*}[t]
    \centering

    \begin{subfigure}[t]{0.32\textwidth}
        \centering
        \includegraphics[width=\linewidth]{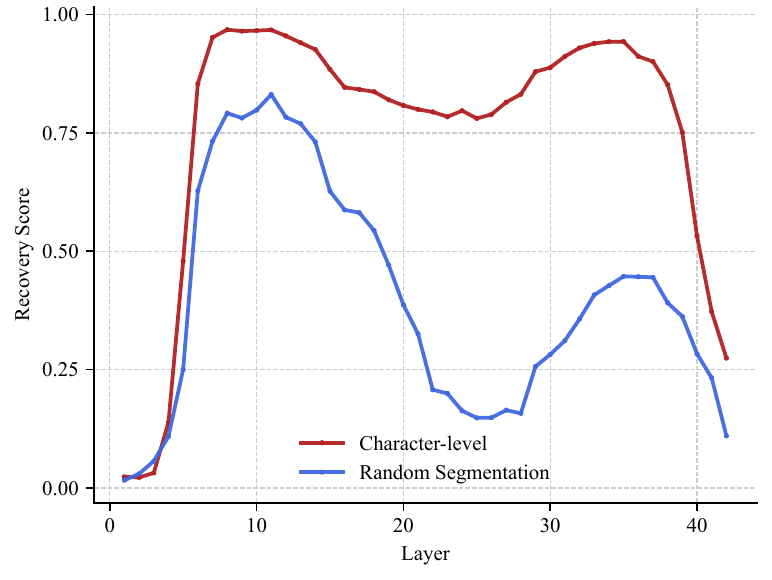}
        \caption{Gemma-2-9B-It}
    \end{subfigure}
    \hfill
    \begin{subfigure}[t]{0.32\textwidth}
        \centering
        \includegraphics[width=\linewidth]{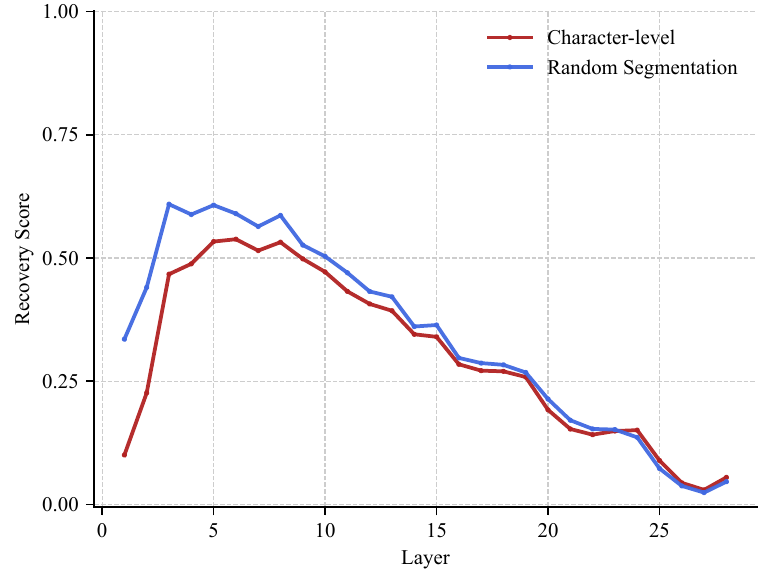}
        \caption{Qwen2.5-7B-Instruct}
    \end{subfigure}
    \hfill
    \begin{subfigure}[t]{0.32\textwidth}
        \centering
        \includegraphics[width=\linewidth]{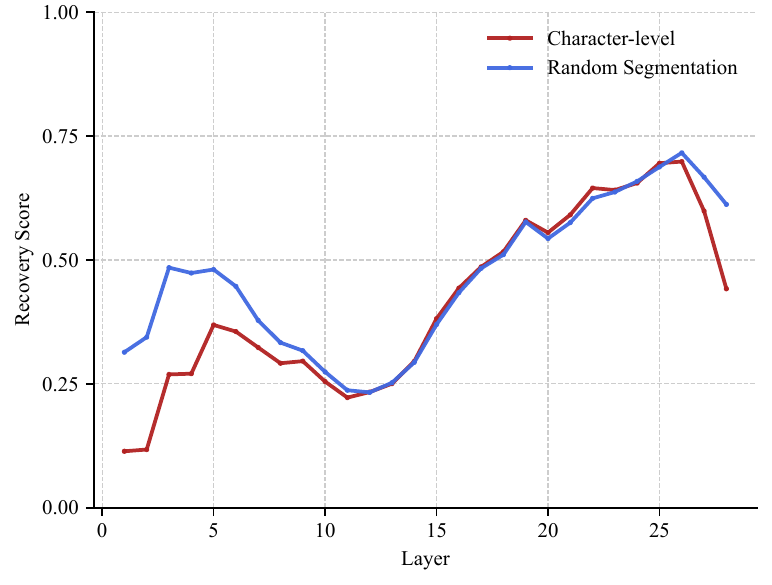}
        \caption{Llama-3.2-3B-Instruct}
    \end{subfigure}

    \caption{Layerwise word recovery under random segmentation tokenization. Recovery dynamics remain qualitatively similar to the character-level setting across all models, suggesting that word recovery generalizes beyond character-level inputs.}
    \label{fig:random_segmentation_appendix}
\end{figure*}

Figure~\ref{fig:random_segmentation_appendix} reports the resulting layerwise recovery curves. Across all models, we observe recovery patterns highly similar to those in the character-level setting. In particular, the same qualitative trends and cross-model differences are preserved: Gemma exhibits strong early-layer recovery, while Qwen and Llama display more gradual or two-stage recovery dynamics. These results suggest that word recovery is not specific to character-level inputs, but instead reflects a more general mechanism underlying robustness to non-canonical tokenization.

\paragraph{Recovery across token lengths.}
We further analyze whether word recovery varies across different categories of canonical tokens. As an initial study, we group tokens by canonical token length (2, 3, 4, and 5+ characters) and measure the maximum recovery score achieved across layers.

Table~\ref{tab:token_length_recovery} summarizes the results. Across all models, longer tokens are generally more difficult to recover, with the effect being most pronounced in Qwen. We hypothesize that recovering longer tokens requires aggregating information across larger character spans through in-group attention, making the process more challenging. Nevertheless, the differences across token lengths remain moderate overall, indicating that word recovery is not driven solely by a small subset of short or easy tokens. Instead, the phenomenon appears to reflect a broad representational mechanism that generalizes across token categories.

\begin{table}[h]
\centering
\small
\begin{tabular}{lcccc}
\toprule
Length & 2 & 3 & 4 & 5+ \\
\midrule
Gemma & 0.83 & 0.90 & 0.91 & 0.82 \\
Qwen & 0.86 & 0.76 & 0.63 & 0.39 \\
Llama & 0.89 & 0.83 & 0.75 & 0.72 \\
\bottomrule
\end{tabular}
\caption{Maximum recovery scores across canonical token lengths. Longer tokens are generally harder to recover, but recovery remains broadly distributed across token categories.}
\label{tab:token_length_recovery}
\end{table}

\begin{figure*}[t]
    \centering

    \begin{subfigure}[t]{0.32\textwidth}
        \centering
        \includegraphics[width=\linewidth]{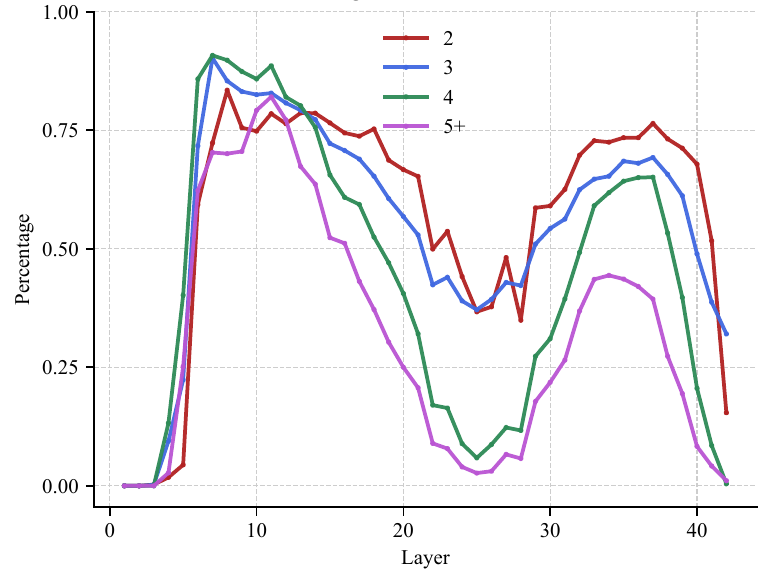}
        \caption{Gemma-2-9B-It}
    \end{subfigure}
    \hfill
    \begin{subfigure}[t]{0.32\textwidth}
        \centering
        \includegraphics[width=\linewidth]{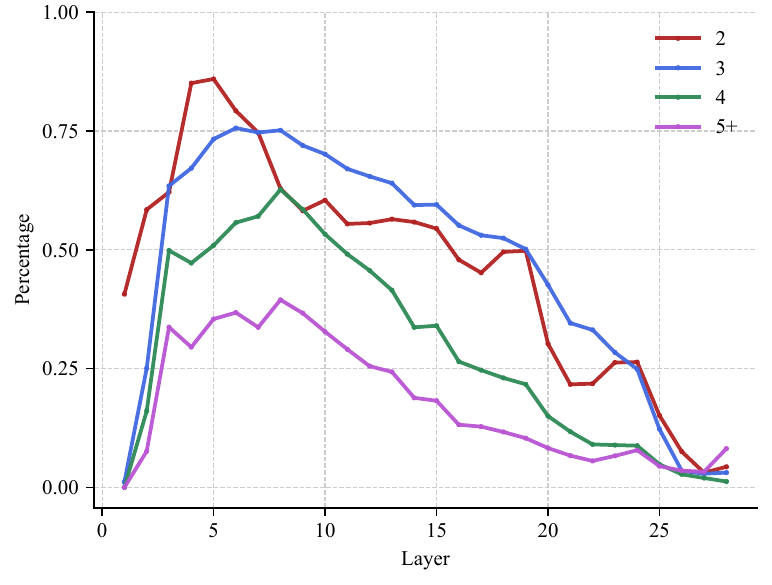}
        \caption{Qwen2.5-7B-Instruct}
    \end{subfigure}
    \hfill
    \begin{subfigure}[t]{0.32\textwidth}
        \centering
        \includegraphics[width=\linewidth]{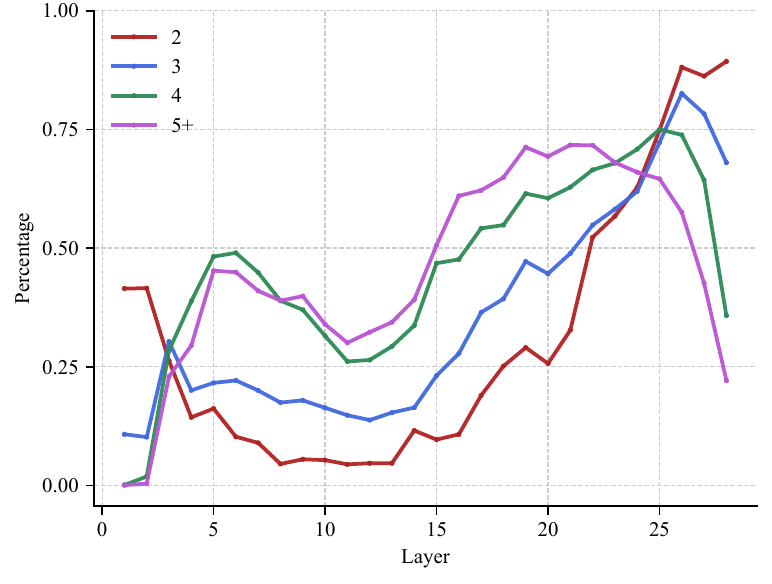}
        \caption{Llama-3.2-3B-Instruct}
    \end{subfigure}

    \caption{Layerwise recovery curves grouped by canonical token length. Longer tokens generally exhibit weaker recovery, though the qualitative recovery dynamics remain consistent across token categories.}
    \label{fig:token_length_appendix}
\end{figure*}

Figure~\ref{fig:token_length_appendix} further reports the full layerwise recovery curves grouped by token length.

\section{Model Size and Architectural Effects on Word Recovery}
\label{sec:appendix_size}
To better understand the disparities in recovery dynamics across model families, we extend our analysis to additional models with different scales, embedding designs, and architectural choices. Specifically, we evaluate Qwen2.5-1.5B-Instruct, Qwen2.5-3B-Instruct, Gemma-2-2B-It, and Mistral-7B-Instruct-v0.1, in addition to the models studied in the main paper. These models allow us to better disentangle the effects of model size, embedding tying, and architectural variation. We first report performance degradation under character-level tokenization on ARC-E in Table~\ref{tab:model_scale}. Across all model families, stronger robustness is generally associated with stronger early-layer recovery.

\begin{table}[h]
\centering
\small
\begin{tabular}{lcc}
\toprule
Model & Canon & Char $\Delta$ \\
\midrule
Qwen2.5-1.5B-Instruct & 89.6 & -24.5 \\
Qwen2.5-3B-Instruct & 92.3 & -6.70 \\
Qwen2.5-7B-Instruct & 95.6 & -2.30 \\
Gemma-2-2B-It & 87.0 & -33.7 \\
Gemma-2-9B-It & 96.8 & -3.50 \\
Mistral-7B-Instruct-v0.1 & 83.5 & -58.4 \\
\bottomrule
\end{tabular}
\caption{Performance degradation under character-level tokenization on ARC-E.}
\label{tab:model_scale}
\end{table}

\paragraph{Qwen family.}
We compare Qwen2.5-1.5B, 3B, and 7B, which share similar training pipelines but differ in both scale and embedding design: the 1.5B and 3B models use tied embeddings, while the 7B model uses untied embeddings. We find that recovery behavior does not align with embedding tying itself. Instead, recovery strength scales consistently with model size: smaller models exhibit weaker early-layer recovery and substantially larger performance degradation, while the 7B model achieves stronger early recovery despite using untied embeddings.

\paragraph{Gemma family.}
We further compare Gemma-2-2B and Gemma-2-9B, which both use tied embeddings but differ in scale. Similar to the Qwen family, the smaller model exhibits weaker early-layer recovery and significantly larger degradation under character-level tokenization. This again suggests that model size strongly correlates with robustness within a model family.

\paragraph{Mistral and architectural effects.}
To probe architectural effects beyond embedding tying, we evaluate Mistral-7B-Instruct-v0.1, which uses sliding-window attention. Despite this architectural difference, Mistral does not exhibit improved recovery behavior relative to other 7B-scale models. Its recovery dynamics are broadly comparable to Qwen-7B, yet its downstream degradation is substantially larger. This suggests that sliding-window attention alone does not explain robustness to character-level tokenization.

\begin{figure*}[t]
    \centering
    
    \begin{subfigure}[t]{0.32\textwidth}
        \centering
        \includegraphics[width=\linewidth]{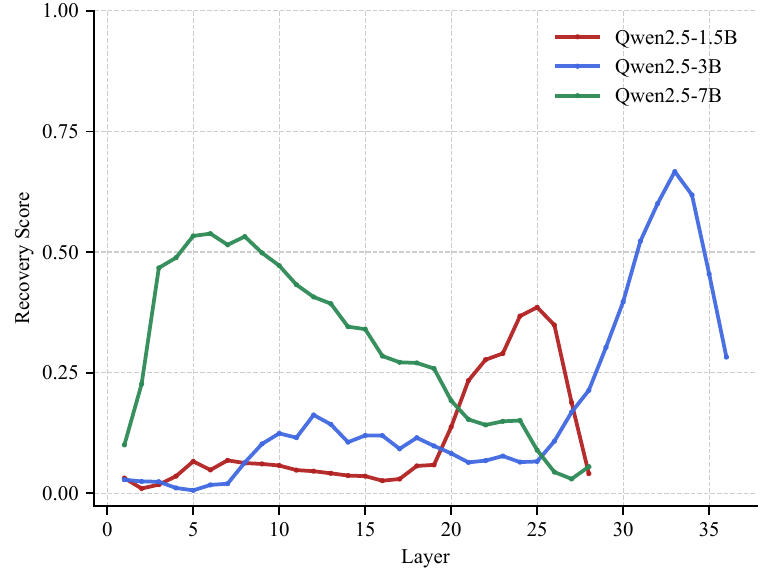}
        \caption{Qwen family}
    \end{subfigure}
    \hfill
    \begin{subfigure}[t]{0.32\textwidth}
        \centering
        \includegraphics[width=\linewidth]{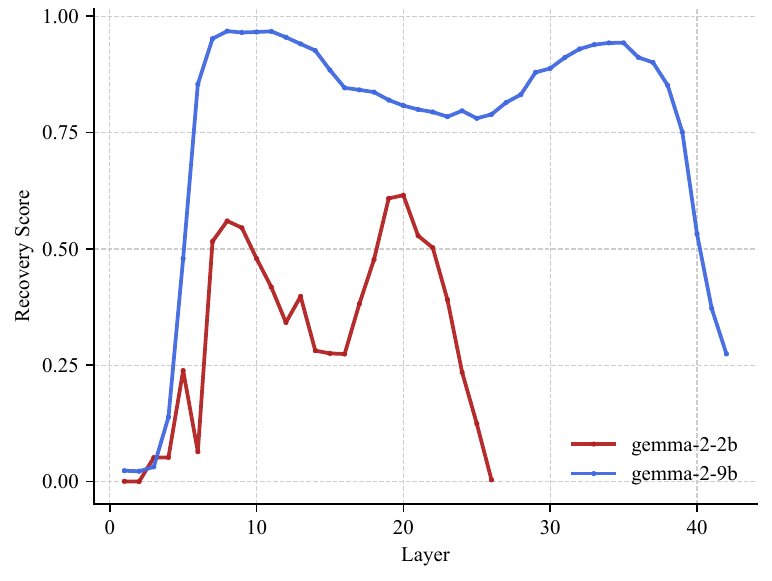}
        \caption{Gemma family}
    \end{subfigure}
    \hfill
    \begin{subfigure}[t]{0.32\textwidth}
        \centering
        \includegraphics[width=\linewidth]{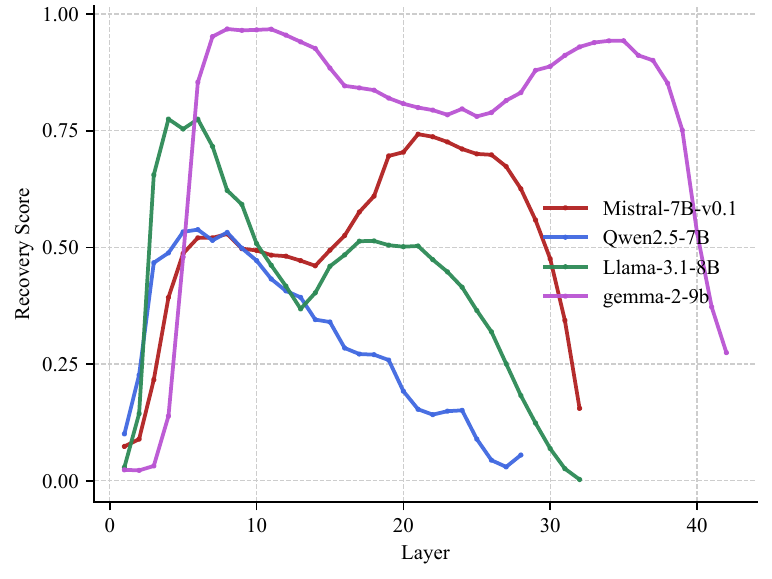}
        \caption{Mistral-7B-Instruct-v0.1}
    \end{subfigure}

    \caption{Layerwise word recovery across additional models under character-level tokenization. (a) Within the Qwen family, recovery strength increases consistently with model size despite differences in embedding tying. (b) Within the Gemma family, the larger model exhibits substantially stronger early-layer recovery and robustness. (c) Mistral-7B-Instruct-v0.1 shows recovery dynamics comparable to other 7B-scale models despite using sliding-window attention, suggesting that this architectural feature alone does not explain robustness.}
    \label{fig:model_architecture_recovery}
\end{figure*}
Figure~\ref{fig:model_architecture_recovery} summarizes the recovery dynamics across these additional models. Overall, our experiments reveal two main findings. First, within a model family, stronger early-layer recovery consistently correlates with improved robustness under character-level tokenization, and model size appears to be an important factor driving this trend. Second, cross-family differences remain difficult to explain through any single architectural component we tested, including embedding tying or sliding-window attention. In particular, models with similar recovery profiles can still exhibit substantially different robustness behaviors. 

\section{Qualitative Validation of In-Group Attention Masking}
\label{sec:appendix_qualitative_masking}
One possible concern regarding the in-group attention masking results is that the intervention may globally disrupt model functionality, rather than specifically impairing word recovery. To investigate this, we conduct a qualitative experiment on the character-level input: \texttt{W h a t \_ i s \_ t h e \_ c a p i t a l \_ o f \_ F r a n c e ?}

We apply in-group attention masking only to the characters corresponding to the canonical token ``France'' (i.e., F--r--a--n--c--e), while leaving all other tokens unchanged. We then compare the model's next-token predictions with and without masking.

Table~\ref{tab:qualitative_masking} reports the top-10 decoding probabilities. Without masking, the model assigns extremely high probability to the correct answer ``Paris'' ($\approx 0.9998$). Under masking, the probability of ``Paris'' drops substantially ($\approx 0.18$), and the model instead distributes probability mass across many alternative capital cities, including ``Rome'', ``Berlin'', ``Madrid'', and ``London''.

Importantly, the model still generates semantically plausible outputs under masking, rather than degenerating into nonsensical predictions. This suggests that the intervention does not globally destroy model functionality. Instead, masking specifically impairs the model's ability to aggregate the characters of ``France'' into a coherent lexical representation, increasing uncertainty about the corresponding entity while preserving general contextual reasoning.

\begin{table*}[t]
\centering
\small
\begin{tabular}{ll}
\toprule
Clean run & Masking (France) \\
\midrule
``Paris'': 0.9997710586 & ``Paris'': 0.1813361049 \\
``paris'': 9.157e-05 & ``Rome'': 0.1692125201 \\
``Par'': 5.018e-05 & ``Berlin'': 0.1167379990 \\
``París'': 3.708e-05 & ``Madrid'': 0.0662391782 \\
\begin{CJK*}{UTF8}{gbsn}``パリ'' \end{CJK*}: 1.445e-05 & ``Toronto'': 0.0468508899 \\
``Pari'': 9.783e-06 & ``Re'': 0.0389084555 \\
``PARIS'': 7.424e-06 & ``Frankfurt'': 0.0383802652 \\
\begin{CJK*}{UTF8}{gbsn}``巴黎'' \end{CJK*}: 6.161e-06 & ``London'': 0.0292352270 \\
``Paris'': 4.248e-06 & ``Nairobi'': 0.0235095415 \\
``P'': 3.472e-06 & ``Helsinki'': 0.0214511715 \\
\bottomrule
\end{tabular}
\caption{Top-10 next-token predictions for the query ``What is the capital of France?'' under character-level tokenization. In-group attention masking is applied only to the characters corresponding to ``France''.}
\label{tab:qualitative_masking}
\end{table*}

\section{Effect of Top-$K$ Choice in Word Recovery Decoding}
\label{sec:appendix_k_choice}

In the main experiments, we measure word recovery by decoding the top-$K$ predicted tokens from each character-level hidden state and aggregating them across the sequence. This choice introduces a potential sensitivity to the value of $K$: smaller $K$ values provide a stricter notion of recoverability, while larger $K$ values may increase recall at the cost of including less confident predictions.

\begin{figure*}[hbp]
    \centering
    \begin{subfigure}[t]{0.25\textwidth}
        \centering
        \includegraphics[width=\textwidth]{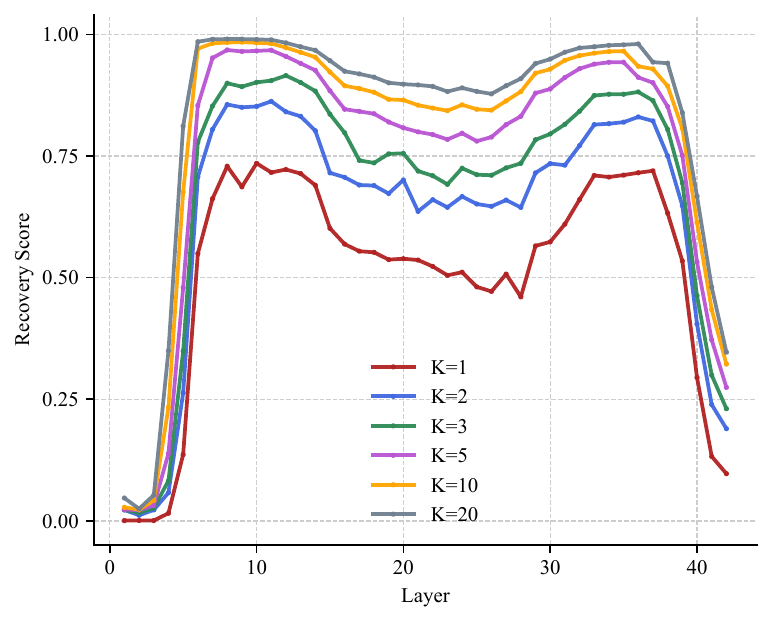}
        \caption{ARC-E}
    \end{subfigure}\hfill
    \begin{subfigure}[t]{0.25\textwidth}
        \centering
        \includegraphics[width=\textwidth]{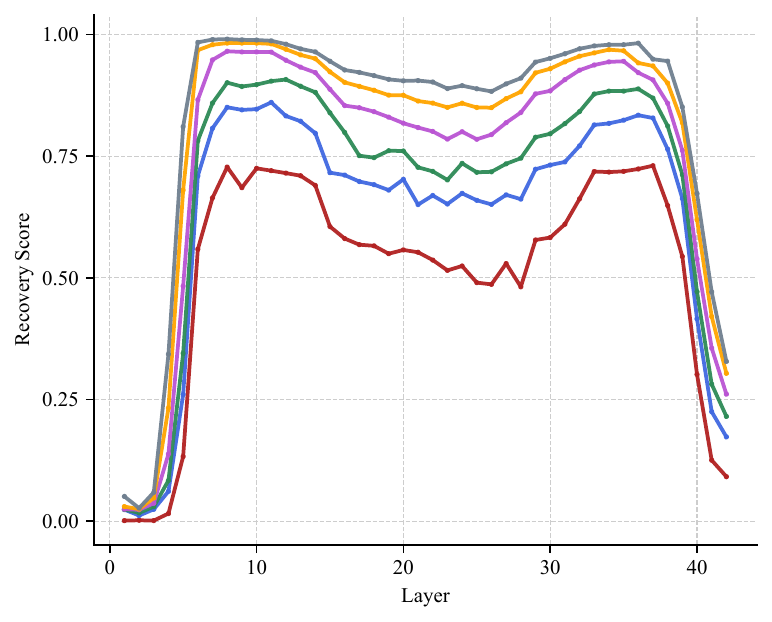}
        \caption{ARC-C}
    \end{subfigure}\hfill
    \begin{subfigure}[t]{0.25\textwidth}
        \centering
        \includegraphics[width=\textwidth]{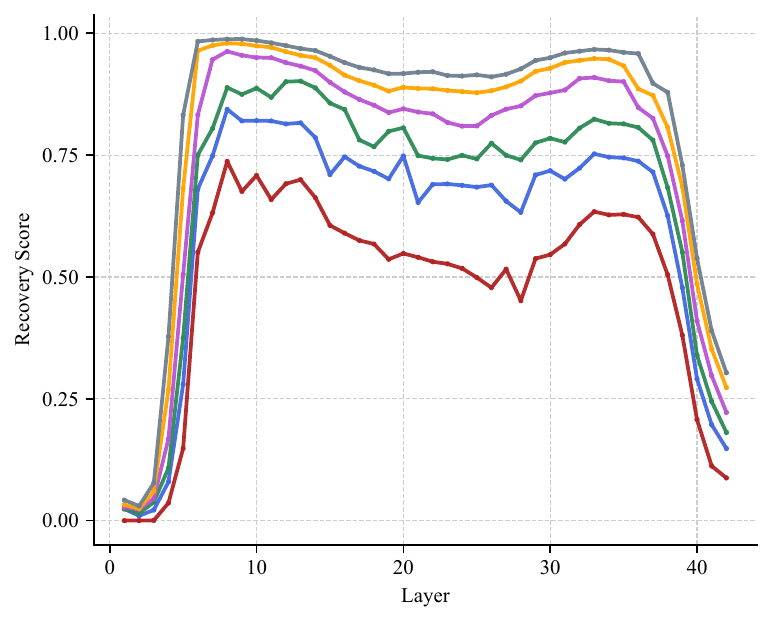}
        \caption{CSQA}
    \end{subfigure}\hfill
    \begin{subfigure}[t]{0.25\textwidth}
        \centering
        \includegraphics[width=\textwidth]{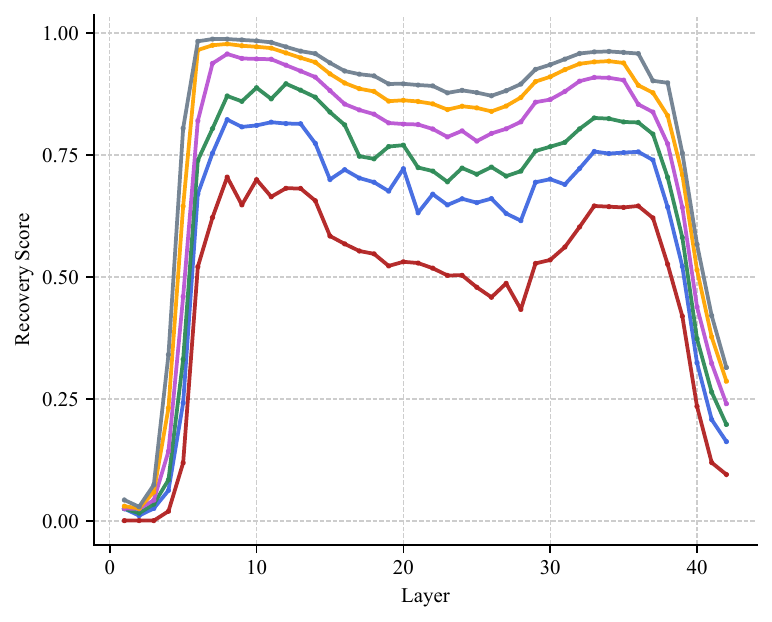}
        \caption{OpenbookQA}
    \end{subfigure}\hfill
\caption{
\textbf{Effect of top-$K$ choice on word recovery for Gemma-2-9B-It.}
We report layerwise word recovery scores under character-level tokenization for different values of $K \in \{1,2,3,5,10,20\}$ across four datasets. While larger $K$ yields uniformly higher absolute recovery scores, the overall layerwise trends and relative recovery dynamics remain consistent across choices of $K$, indicating that our findings are not sensitive to the specific value of $K$.}

    \label{fig:topk_gemma}
\end{figure*}

To assess the robustness of our findings to this design choice, we repeat the word recovery analysis using a range of values $K \in \{1, 2, 3, 5, 10, 20\}$. Across all datasets, we find that the qualitative behavior of word recovery remains consistent (Figure~\ref{fig:topk_gemma}). In particular, the layerwise recovery curves exhibit similar shapes, and the same early-vs.-late recovery patterns described in the main text persist for all tested values of $K$.

While increasing $K$ naturally leads to higher absolute recovery scores, the onset layer at which recovery begins and the saturation behavior across layers are largely unchanged. This indicates that our conclusions about the emergence and dynamics of word recovery do not depend on a specific choice of $K$. Based on this analysis, we use $K=5$ in the main experiments as a balance between strictness and robustness.

\section{Additional Analysis: In-Group Word Recovery Decoding}
\label{sec:appendix_ingroup_decoding}

In the main analysis, we define word recovery using a set-based decoding formulation that aggregates top-$K$ predictions across the entire character sequence (Section~\ref{sec:word_recovery}). While this formulation captures whether canonical lexical identities are present anywhere in the hidden states, it may overestimate recovery if a token is predicted spuriously at positions unrelated to its original character span. To address this concern, we conduct a more restrictive decoding analysis that limits recovery to \emph{in-group character ranges}. Specifically, for each canonical token $t_i$ with character span $\mathcal{G}(t_i) = [s_i, e_i)$, we restrict decoding to hidden states within this span. Let
\[
\mathcal{P}^{(\ell)}_{\text{group}}(t_i) = \bigcup_{j=s_i}^{e_i-1} \mathrm{TopK}(p^{(\ell)}_j),
\]
and we say that $t_i$ is recovered at layer $\ell$ under in-group decoding if $t_i \in \mathcal{P}^{(\ell)}_{\text{group}}(t_i)$. The corresponding in-group recovery score is defined as
\[
R^{(\ell)}_{\text{group}} = \frac{1}{|\mathcal{T}|} \sum_{t_i \in \mathcal{T}} \mathbb{I}\big[t_i \in \mathcal{P}^{(\ell)}_{\text{group}}(t_i)\big].
\]

This formulation ensures that a canonical token is counted as recovered only if its identity is decodable from the hidden states of its own character span, rather than from unrelated positions elsewhere in the sequence.

\begin{figure*}[hbp]
    \centering
    \begin{subfigure}[t]{0.32\textwidth}
        \centering
        \includegraphics[width=\textwidth]{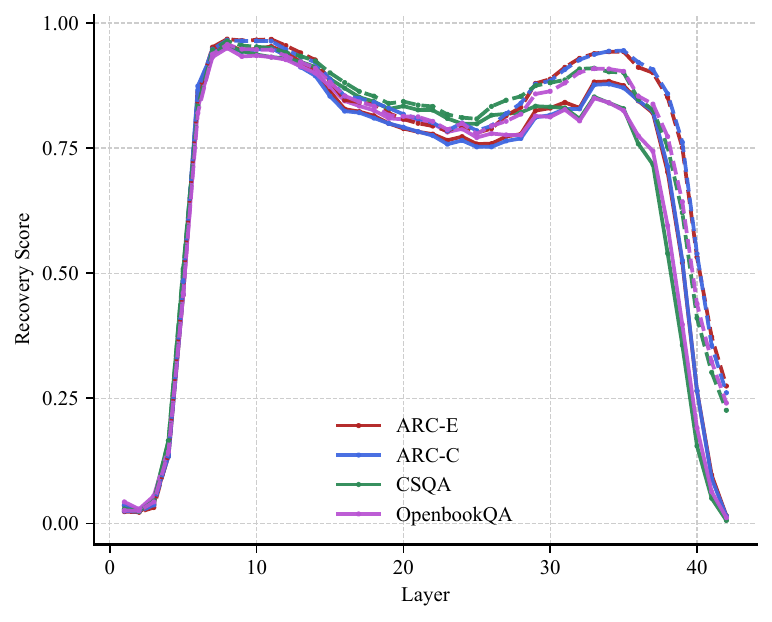}
        \caption{Gemma-2-9b-It}
    \end{subfigure}\hfill
    \begin{subfigure}[t]{0.32\textwidth}
        \centering
        \includegraphics[width=\textwidth]{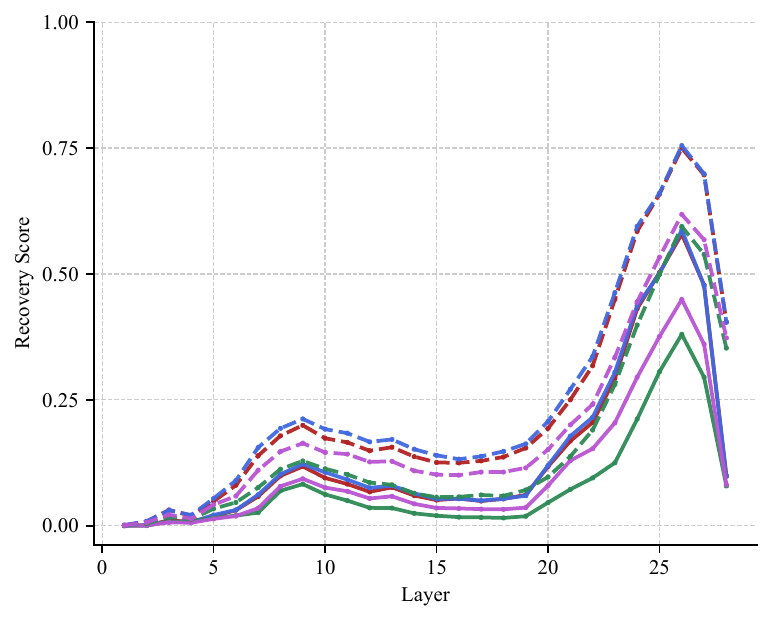}
        \caption{Qwen2.5-7B-Instruct}
    \end{subfigure}\hfill
    \begin{subfigure}[t]{0.32\textwidth}
        \centering
        \includegraphics[width=\textwidth]{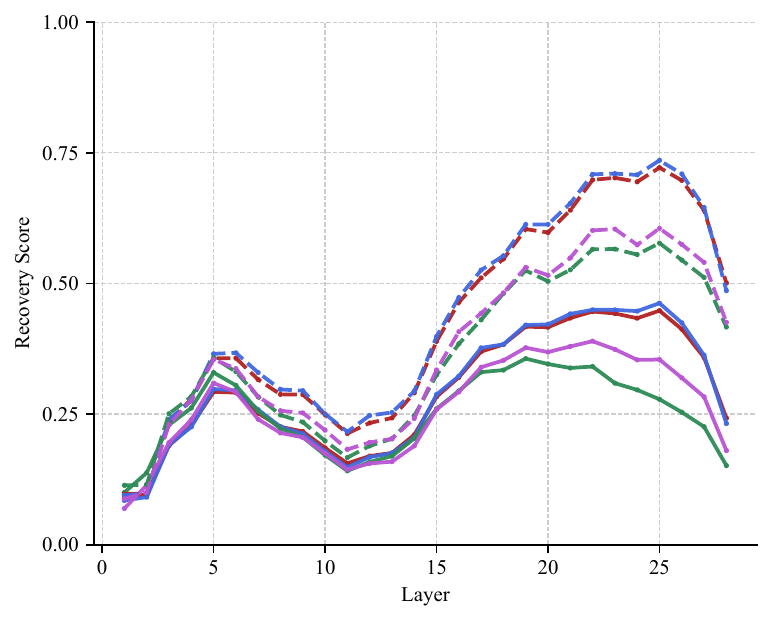}
        \caption{Llama-3.2-3B-Instruct}
    \end{subfigure}\hfill
\caption{
\textbf{In-group word recovery decoding across layers.}
We report word recovery scores when decoding is restricted to the character span corresponding to each canonical token (in-group decoding). Dashed lines show the original set-based decoding over the full character sequence. While in-group decoding yields lower absolute recovery scores, the layerwise recovery dynamics and relative trends remain consistent, indicating that word recovery does not arise from spurious matches elsewhere in the sequence.}

    \label{fig:group_decoding}
\end{figure*}

Across all models and datasets, we find that in-group decoding yields slightly lower absolute recovery scores than the set-based formulation, as expected under the stricter criterion (Figure~\ref{fig:group_decoding}). However, the qualitative conclusions remain unchanged. In particular, (i) word recovery still emerges reliably across layers under character-level tokenization, (ii) the relative ordering of models and datasets is preserved, and (iii) the layerwise recovery dynamics, early saturation in Gemma-2-9B-It and two-stage recovery in Llama-3.2-3B-Instruct, remain consistent.

These results indicate that the main findings are not driven by spurious global decoding effects. Instead, recovered word-level representations are localized to the character spans from which they originate, supporting the interpretation of word recovery as a genuine internal aggregation process rather than an artifact of set-based decoding.

\end{document}